\documentclass[runningheads]{llncs}
\usepackage{graphicx}

\usepackage{tikz}
\usepackage{comment} 
\usepackage{amsmath,amssymb} 
\usepackage{color}

\usepackage{epsfig}
\usepackage{graphicx}
\usepackage{amsmath}
\usepackage{amssymb}
\usepackage{array}
\usepackage{cite}

\usepackage{algorithm}
\usepackage{algorithmic}
\usepackage{mdwmath}

\usepackage{graphicx}         
\usepackage{subfigure}
\usepackage{pifont}

\usepackage{bbm}
\usepackage{dsfont}

\usepackage{bm}
\usepackage{comment}
\usepackage{amsmath}
\usepackage{enumerate}  
\usepackage{mathrsfs}
\usepackage{makeidx}         
\usepackage{graphicx}        
\usepackage{subfigure}
\usepackage{epsfig,amssymb,latexsym}
\usepackage{psfrag}
\usepackage{algorithmic}
\usepackage{algorithm}
\usepackage{rotating}
\usepackage[misc]{ifsym}

\usepackage{fancyhdr}
\usepackage{multirow}

\usepackage{color}
\usepackage{indentfirst}

\renewcommand{\algorithmicrequire}{\textbf{Input:}}   
\renewcommand{\algorithmicensure}{\textbf{Output:}}   

\newcommand{\mr}{\mathrm}

\usepackage{colortbl}
\definecolor{lightgray}{rgb}{.93,.93,.93}

\begin{document}
\pagestyle{headings}
\mainmatter
\def\ECCVSubNumber{696}  

\title{CSCL: Critical Semantic-Consistent Learning for Unsupervised Domain Adaptation} 

\titlerunning{CSCL: Critical Semantic-Consistent Learning}
%
\author{
Jiahua Dong\inst{1,2,3~}\orcidID{0000-0001-8545-4447} \and
Yang Cong\inst{1,2,}\thanks{The corresponding author is Prof. Yang Cong.} \and
Gan Sun\inst{1,2,}\thanks{The author contributes equally to this work.} \and \\
Yuyang Liu\inst{1,2,3} \and 
Xiaowei Xu\inst{4}
}
\authorrunning{J. Dong et al.}
%
\institute{
State Key Laboratory of Robotics, Shenyang Institute of Automation, \\ Chinese Academy of Sciences, Shenyang, 110016, China. \and
Institutes for Robotics and Intelligent Manufacturing, \\ 
Chinese Academy of Sciences, Shenyang, 110016, China.  \and
University of Chinese Academy of Sciences, Beijing, 100049, China. \and
Department of Information Science, University of Arkansas at Little Rock, USA. \\
\email{\{dongjiahua, liuyuyang\}@sia.cn, \{congyang81, sungan1412\}@gmail.com, xwxu@ualr.edu}
}
\maketitle

\begin{abstract}
Unsupervised domain adaptation without consuming annotation process for unlabeled target data attracts appealing interests in semantic segmentation. However, 1) existing methods neglect that not all semantic representations across domains are transferable, which cripples domain-wise transfer with untransferable knowledge; 2) they fail to narrow category-wise distribution shift due to category-agnostic feature alignment. To address above challenges, we develop a new \underline{C}ritical \underline{S}emantic-\underline{C}onsistent \underline{L}earning (CSCL) model, which mitigates the discrepancy of both domain-wise and category-wise distributions. Specifically, a critical transfer based adversarial framework is designed to highlight transferable domain-wise knowledge while neglecting untransferable knowledge. Transferability-critic guides transferability-quantizer to maximize positive transfer gain under reinforcement learning manner, although negative transfer of untransferable knowledge occurs. Meanwhile, with the help of confidence-guided pseudo labels generator of target samples, a symmetric soft divergence loss is presented to explore inter-class relationships and facilitate category-wise distribution alignment. Experiments on several datasets demonstrate the superiority of our model.

\end{abstract}

\section{Introduction}
Convolutional neural networks relying on a large amount of annotations have achieved significant successes in many computer vision tasks, \emph{e.g.}, semantic segmentation \cite{net:deeplab, Zhu_2019_ICCV, Shelhamer:2017:FCN:3069214.3069246}. Unfortunately, the learned models could not generalize well to the unlabeled target domain, especially when there is a large distribution gap between the training and evaluation datasets. Unsupervised domain adaptation \cite{exp:LtA, Saito_2018_CVPR, Zou_2018_ECCV, Dong_2020_CVPR} shows appealing segmentation performance for unlabeled target domain by transferring effective domain-invariant knowledge from labeled source domain. To this end, enormous related state-of-the-art models \cite{Vu_2019_CVPR, Li_2019_CVPR, Tsai_2019_ICCV, Luo_2019_ICCV, Ding_2018_ECCV} are developed to mitigate the distribution discrepancy between different datasets.

\begin{figure*}[t]
	\small
	\centering
	\includegraphics[trim = 4mm 9mm 0mm 0mm 5mm, clip, width =352pt, height =115pt]{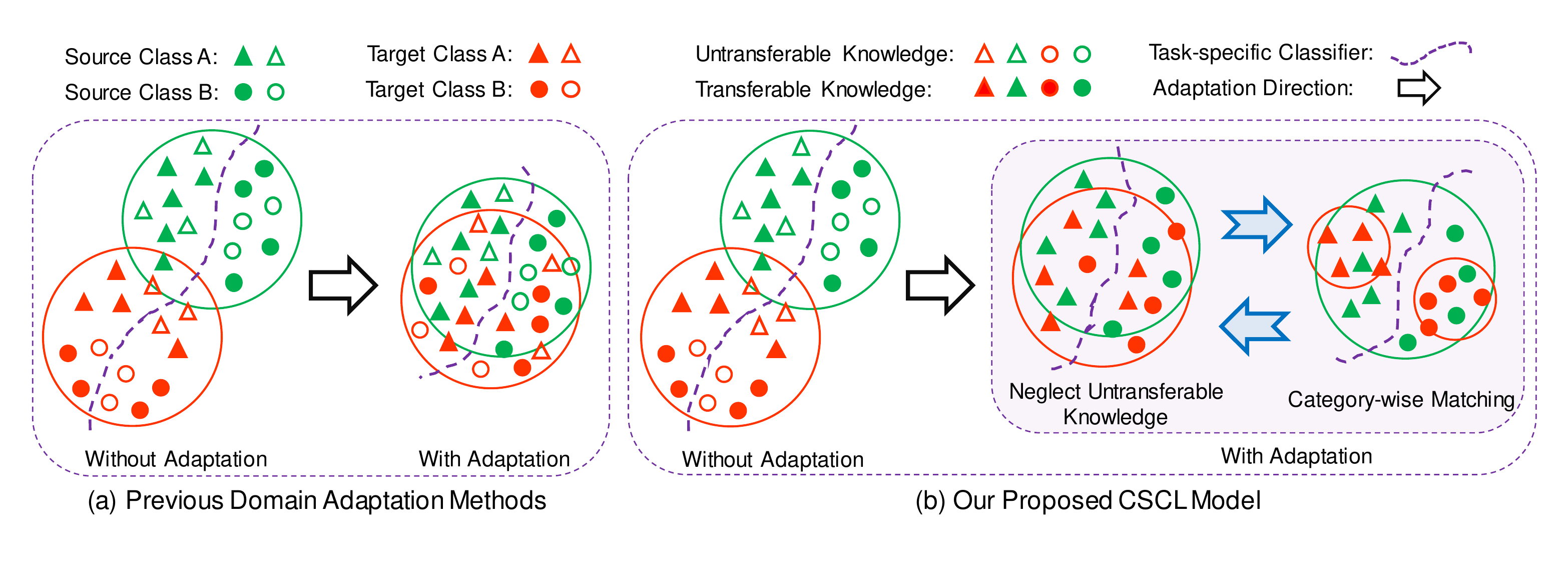}
	\caption{Illustration of previous domain adaptation methods and our proposed CSCL model. (a): Existing models neglect the negative transfer brought by untransferable knowledge, and result in the category-wise distribution mismatch due to the lack of valid labels of target samples. (b): As for our CSCL model, a critical transfer based adversarial framework is developed to prevent the negative transfer of untransferable knowledge. Moreover, a symmetric soft divergence loss is designed to align category-wise distributions with the assistance of a confidence-guided pseudo labels generator.} 
	\label{fig:illustration_of_our_model} 
\end{figure*}

However, most existing methods \cite{Dong_2019_ICCV, Lian_2019_ICCV, Li_2019_CVPR, Tsai_2019_ICCV, Luo_2019_ICCV, Gong_2019_CVPR} ignore the fact that not all semantic representations among source and target datasets are transferable, while forcefully taking advantage of untransferable knowledge leads to negative domain-wise transfer, as shown in Fig.~\ref{fig:illustration_of_our_model} (a). In other words, semantic representations across domains cannot contribute equally to narrowing domain-wise distribution shift. For example, unbalanced object categories and objects with various appearances in different datasets are not equally essential to facilitate the semantic transfer. Moreover, category-wise distributions cannot be matched well across domains \cite{exp:LtA, exp:CCA, exp:LSD, exp:CGAN, Wu_2018_ECCV, Saito_2018_CVPR, Xia_2020_CVPR} since the lack of valid labels of target samples results in the category-agnostic feature alignment. As depicted in Fig.~\ref{fig:illustration_of_our_model} (a), the semantic features of same class among different datasets are not mapped nearby. Therefore, exploring efficient semantic knowledge to narrow both domain-wise and category-wise distributions discrepancy is a crucial challenge.

To tackle above mentioned challenges, as presented in Fig.~\ref{fig:illustration_of_our_model} (b), a new \underline{C}ritical \underline{S}emantic-\underline{C}onsistent \underline{L}earning (CSCL) model is developed for unsupervised domain adaptation. \textbf{1)} On one hand, we design a critical transfer based adversarial framework to facilitate the exploration of transferable domain-wise representations while preventing the negative transfer of untransferable knowledge. To be specific, the transferability-quantizer highlights domain-wise semantic representations with high transfer scores, and transferability-critic evaluates the quality of corresponding transfer process. With a direct supervision from the transferability-critic, positive transfer feedback guides the transferability-quantizer to maximize the transfer gain, even though negative transfer of untransferable knowledge occurs. Both the transferability-critic and transferability-quantizer are trained under the reinforcement learning manner to narrow the marginal distribution gap. 
\textbf{2)} On the other hand, confident soft pseudo labels for target samples are progressively produced by the confidence-guided pseudo label generator, which efficiently attenuates the misleading effect brought by incorrect or ambiguous pseudo labels. With the guidance of generated soft pseudo labels, a symmetric soft divergence loss is developed to explore inter-class relationships across domains, and further bridge category-wise conditional distribution shift across domains. 
Therefore, our proposed CSCL model could not only learn discriminative transferable knowledge with high transfer scores for target samples prediction, but also well match both domain-wise and category-wise semantic consistence by minimizing marginal and conditional distributions gaps. Experiments on several datasets illustrate the effectiveness of our model with state-of-the-art performance.

The main contributions of this work are summarized as follows:
\begin{itemize}
	\setlength{\itemsep}{3pt}		
	\setlength{\parsep}{0pt}	
	\setlength{\parskip}{0pt}
	\item A new Critical Semantic-Consistent Learning (CSCL) model is proposed to facilitate the exploration of both domain-wise and category-wise semantic consistence for unsupervised domain adaptation. To our best knowledge, this is the first attempt to highlight transferable knowledge via a critical transfer mechanism while neglecting untransferable representations. 
	
	\item A critical transfer based adversarial framework is developed to explore transferable knowledge via a transferability-quantizer, while preventing the negative transfer of irrelevant knowledge via a transferability-critic.   
	
	\item A symmetric soft divergence loss is designed to explore inter-class relationships and narrow category-wise distribution shift, under the supervision of confident pseudo labels mined by confidence-guided pseudo labels generator.
	
\end{itemize}

\section{Related Work}
\textbf{Semantic Segmentation:}
Deep neural network \cite{wang2019laplacian, wang2020recurrent, zhang2019visual} based semantic segmentation has caught growing research attention recently. \cite{Shelhamer:2017:FCN:3069214.3069246} develops the first fully convolutional network for pixel-level prediction. To enlarge the receptive field, \cite{Chen_Semantic2014, Yu2015MultiScaleCA} propose the dilated convolution operator. Later, encoder-decoder networks, such as U-net \cite{RonnebergerFB15}, Deeplab \cite{net:deeplab}, etc, are proposed to fuse low-level and high-level information. \cite{Zhao_2018_ECCV, NIPS2017_6750, Liu_2015_ICCV} aim to capture long-range contextual dependencies. However, they require large-scale labeled samples to attain remarkable performance, which is time-consuming to manually annotate the data.

\textbf{Unsupervised Domain Adaptation:}
After generative adversarial network \cite{Goodfellow:2014:GAN} is first employed by Hoffman \emph{et al.} \cite{exp:Wild} to narrow distribution discrepancy for domain adaptation, enormous adversarial learning based variants \cite{exp:LtA, exp:CCA, exp:LSD, exp:CGAN, Wu_2018_ECCV, Saito_2018_CVPR, Dong_2019_ICCV, Luo_2019_ICCV, Tsai_2019_ICCV, Dong_2020_CVPR} are proposed to learn domain-invariant knowledge with a domain classifier. Another important strategy is curriculum learning \cite{exp:CL, Lian_2019_ICCV}, which predicts target labels according to source distribution properties. \cite{Zou_2018_ECCV, Zou_2019_ICCV} develops a self-training non-adversarial model to mitigate the discrepancy shift with the assistance of pseudo labels. Li \emph{et al.} \cite{Li_2019_CVPR} design a bidirectional learning model for knowledge transfer, which incorporates appearance translation module and semantic adaptation module. \cite{Gong_2019_CVPR} considers intermediate optimal transfer domain for semantic segmentation. Moreover, some novel discriminative losses, such as sliced Wasserstein discrepancy \cite{Lee_2019_CVPR}, adversarial entropy minimization \cite{Vu_2019_CVPR} and adaptive category-level adversarial loss \cite{Luo_2019_CVPR}, are proposed to minimize distribution discrepancy between different datasets.

\begin{figure*}[t]
	\small
	\centering
	\includegraphics[trim = 5mm 8mm 0mm 5mm, clip, width =355pt, height =107pt]{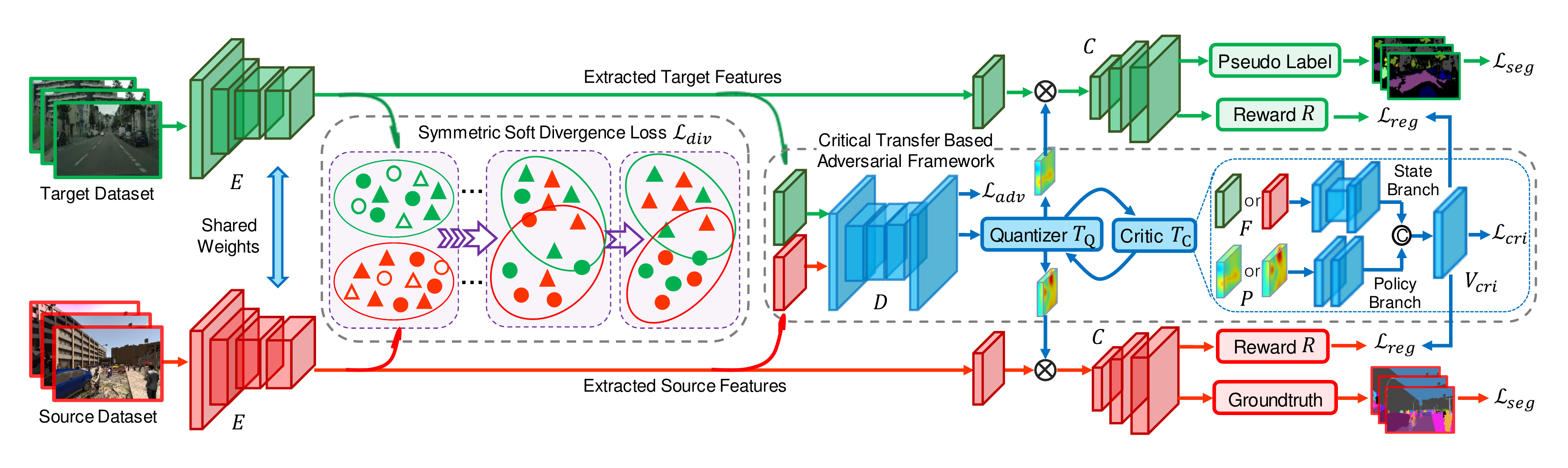}
	\caption{Overview of our CSCL model. A critical transfer based adversarial framework is developed to highlight transferable knowledge while preventing the negative transfer of untransferable knowledge, where  $T_C$ guides $T_Q$ to maximize the transfer gain. A symmetric soft divergence loss $\mathcal{L}_{div}$ aims to explore inter-class relationships and align category-wise distribution with the help of confidence-guided pseudo labels generator.} 
	\label{fig:overview_of_our_model}  
\end{figure*}

\section{The Proposed CSCL Model}  

\subsection{Overview} 
Given a labeled source dataset $D_s = \{x_i^s, y_i^s\}_{i=1}^{n_s}$ with $n_s$ samples, where $x_i^s$ and $y_i^s$ denote the image and corresponding pixel label, respectively. Define $D_t = \{x_j^t\}_{j=1}^{n_t}$ as the unlabeled target domain with $n_t$ images. In unsupervised domain adaptation task, source and target domains share the consistent annotations. Our goal is to explore the transferable knowledge across domains for classifying unlabeled target data, even though there is a large distribution discrepancy.

The overview of our proposed CSCL model is depicted in Fig.~\ref{fig:overview_of_our_model}. The source images $x_i^s$ with annotations $y_i^s$ are first forwarded into the segmentation network to optimize the encoder $E$ and pixel classifier $C$ via the loss $\mathcal{L}_{seg}$. Then the semantic features of both source and target samples extracted by encoder $E$ are employed to update discriminator $D$ via adversarial objective $\mathcal{L}_{adv}$. Unfortunately, such strategy cannot align domain-wise distribution well to some degree, since not all semantic knowledge across domains are transferable. Untransferable knowledge significantly degrades the transfer performance, which is obviously neglected by most existing models \cite{Lee_2019_CVPR, Vu_2019_CVPR, Luo_2019_CVPR, Tsai_2019_ICCV, Luo_2019_ICCV, Lian_2019_ICCV}. In light of this issue, a critical transfer based adversarial framework in Section \ref{sec:transferability_critic_framework} is developed to explore transferable knowledge while neglecting irrelevant knowledge. Furthermore, category-wise distribution shift is difficult to narrow due to lack of valid pixel labels of target samples. Therefore, a symmetric soft divergence loss $\mathcal{L}_{div}$ in Section \ref{sec:soft_divergence_loss} is designed to facilitate category-wise distribution alignment, with the assistance of pseudo labels generator in Section \ref{sec:pseudo_labels_generator}.

\subsection{Critical Transfer Based Adversarial Framework}  \label{sec:transferability_critic_framework} 
Adversarial learning is utilized to encourage semantic features among different datasets share consistent domain-wise distribution, as presented in Fig.~\ref{fig:overview_of_our_model}. The features extracted from source and target samples are forwarded into discriminator $D$ for distinguishing which one is from the source or target. Meanwhile, encoder $E$ aims to fool $D$ by encouraging the target features to look like the source one. Thus, the generative adversarial objective $\mathcal{L}_{adv}$ is formulated as: 
\begin{equation}
\begin{split}
\min_{\theta_{E}}\max_{\theta_{D}}\mathcal{L}_{adv} = ~& \mathbb{E}_{x_{i}^s \in D_s} \big[1 - \mr{log}(D(E(x_i^s; \theta_E); \theta_{D})) \big] + \\ &\mathbb{E}_{x_{j}^t \in D_t}\big[\mr{\log} (D(E(x_j^t; \theta_E); \theta_{D}))\big], 
\end{split}
\label{equ:adversarial_learning_loss}  	
\end{equation}
where $\theta_{D}$ and $\theta_{E}$ are the parameters of $D$ and $E$. $D(E(x_i^s; \theta_E); \theta_{D})$ and $E(x_i^s; \theta_E)$ denote the probability output of $D$ and the features extracted by $E$ for source sample $x_i^s$, respectively. Furthermore, the definitions of $D(E(x_j^t; \theta_E); \theta_{D})$ and $E(x_j^t; \theta_E)$ are similar to $D(E(x_i^s; \theta_E); \theta_{D})$ and $E(x_i^s; \theta_E)$ as well.

However, Eq.~\eqref{equ:adversarial_learning_loss} cannot efficiently align domain-wise distributions between different datasets, since not all semantic representations across domains are transferable. To prevent the negative transfer brought by untransferable knowledge, we develop a critical transfer based adversarial framework, as depicted in Fig.~\ref{fig:overview_of_our_model}. Specifically, transferability-quantizer $T_Q$ highlights transferable knowledge while neglecting untransferable knowledge. Transferability-critic $T_C$ examines the quality of quantified transferability via $T_Q$, and feedbacks positive supervision to guide $T_Q$ when negative transfer of untransferable knowledge occurs. The details about our critical transfer based adversarial framework are as follows:

\textbf{Transferability-Quantizer $T_Q$:} 
According to the capacity of discriminator $D$ in identifying whether the input is from source or target datasets, we can easily distinguish which representations across domains are transferable, already transferred or untransferable. In other words, the output probabilities of $D$ can determine whether the corresponding features are transferable or not for domain adaptation. For an intuitive example, the features that are already adapted across different datasets will confuse $D$ to distinguish whether the input is from the source or target. Consequently, $T_Q$ encodes the output probability of $D$ via one convolutional layer to highlight the relevance transferability of transferable knowledge, and then utilizes uncertainty measure function of information theory $\mathcal{U}(f) = -\sum_b f_b \mathrm{log}(f_b)$ to quantify the transferability of corresponding semantic features. Generally, the quantified transferability $P_i^s$ and $P_j^t$ for source image $x_i^s$ and target sample $x_j^t$ are respectively defined as:
\begin{equation} 
P_i^s = 1 - \mathcal{U}\big(T_Q(D(E(x_i^s; \theta_E); \theta_{D}); \theta_{T_Q})\big); P_j^t = 1 - \mathcal{U}\big(T_Q(D(E(x_j^t; \theta_E); \theta_{D});\theta_{T_Q})\big), 
\label{eq:quantified_transferability} 
\end{equation} 
where $T_Q(D(E(x_i^s; \theta_E); \theta_{D}); \theta_{T_Q})$ and $T_Q(D(E(x_j^t; \theta_E); \theta_{D});\theta_{T_Q})$ are the outputs of $T_Q$ with network parameters $\theta_{T_Q}$ for source and target samples.
Since the false or misleading transferability highlighted by $T_Q$ may degrade transfer performance to some degree, transferability-critic $T_C$ is developed to evaluate the transfer quality of $T_Q$ and maximize the positive transfer gain.

\textbf{Transferability-Critic $T_C$:} 
Due to the non-differentiability of most evaluation strategies (\emph{e.g.}, the transfer gain), transferability-critic $T_C$ is trained under the reinforcement learning manner. Specifically, the input sample $x$ ($x_i^s$ or $x_j^t$) is regarded as the state in each training step, and the extracted feature $F$ via encoder $E$ is formulated as $F = E(x; \theta_E)$. Afterwards, based on the input feature $F$, $T_Q$ predicts the quantified transferability $P$ ($P_i^s$ or $P_j^t$) for sample $x$, \emph{i.e.}, $P = 1 - \mathcal{U}\big(T_Q(D(E(x; \theta_E); \theta_{D}); \theta_{T_Q})\big)$. To conduct the maximum transfer gain and evaluate the quality of transferability, the transferability-critic $T_C$ takes both $F$ and $P$ as the inputs, as depicted in Fig.~\ref{fig:overview_of_our_model}. It contains two branches, where the state branch utilizes three convolutional block to extract state information, and the policy branch applies two convolutional layers to encode relevance transferability of transferable representations. Afterwards, the outputs of both state and policy branches are concatenated, which are forwarded into a convolution layer to evaluate the critic value $V_{cri} = T_C(F, P; \theta_{T_C})$, where $\theta_{T_C}$ denotes the network weights of $T_C$. For transferability-quantizer $T_Q$, $\mathcal{L}_{cri}$ is proposed to maximize the transfer gain (\emph{i.e.}, critic value $V_{cri}$), which is defined as follows:
\begin{equation} 
\min_{\theta_{T_Q}}\mathcal{L}_{cri} = \mathbb{E}_{x\in(D_s, D_t)}[-V_{cri}] = \mathbb{E}_{x\in(D_s, D_t)}[-T_C(F, P; \theta_{T_C})].  
\label{eq:critic_loss} 
\end{equation}
Intuitively, $\mathcal{L}_{cri}$ encourages $T_Q$ to highlight knowledge with high transfer scores by generating higher critic value $V_{cri}$, which maximizes the positive transfer gain.

To guide $T_C$ feedback positive transfer gain, a new reward mechanism $R$ is developed, which consists of the segmentation reward $R^s$ and the amelioration reward $R^a$. To be specific, $R^s$ measures whether the transferability quantified via $T_Q$ leads to the correct prediction. The value of $R^s$ at the $n$-th pixel is:
\begin{equation}
\begin{split}
R_n^s =
\left\{
\begin{aligned}		
&1, \quad \mr{if}~\mr{argmax}\big(C(F; \theta_C)_n \big) = \mr{argmax}(y_n), \\
&0, \qquad\qquad\qquad \mr{otherwise},  \\
\end{aligned} 					
\right.											
\end{split}							
\label{equation:definition_R^s}	
\end{equation}
where $\theta_{C}$ denotes the parameters of classifier $C$. $C(F; \theta_C)_n$ denotes the probability outputs at the $n$-th pixel in sample $x$ ($x_i^s$ or $x_j^t$). $y_n$ is the one-hot groundtruth $y_i^s$ of $x_i^s$ or soft pseudo label $\hat{y}_j^t$ of $x_j^t$ mined in Section \ref{sec:pseudo_labels_generator} at the $n$-th pixel. Moreover, the amelioration reward $R^a$ examines whether the transferability quantified via $T_Q$ facilitates positive transfer, where the value of $R^a$ at the $n$-th pixel is:
\begin{equation}
\begin{split}
R_n^a =
\left\{
\begin{aligned}		
&1, ~~ \mr{if} ~C(F; \theta_C)_n^k ~ \textgreater~ C^{b}(F; \theta_C)_n^k~\mr{and}~k = \mr{argmax}(y_n) \\
&0, \qquad\qquad\qquad \mr{otherwise}, \\
\end{aligned} 					
\right.											
\end{split}							
\label{equation:definition_R^a}	
\end{equation}
where $C(F; \theta_C)_n^k$ and $C^b(F; \theta_C)_n^k$ respectively represent the probability of the $k$-th class predicted by $C$ with quantified transferability $P$ or not. The overall reward is represented as $R=R^s + R^a$. Thus, to conduct $T_C$ feedback accurate supervision, we develop a reward regression loss $\mathcal{L}_{reg}$, which minimizes the gap between the estimated critic value $V_{cri}$ and the defined reward $R$, \emph{i.e.},
\begin{equation} 
\min_{\theta_{T_C}}\mathcal{L}_{reg} = \mathbb{E}_{x\in(D_s, D_t)}\big[(V_{cri} - R)^2] = \mathbb{E}_{x\in(D_s, D_t)}[\big(T_C(F, P; \theta_{T_C})- R\big)^2\big]. 
\label{eq:reward_regression_loss} 
\end{equation}

\subsection{Confidence-Guided Pseudo Labels Generator}
\label{sec:pseudo_labels_generator} 
To guide the critical transfer based adversarial framework in Section \ref{sec:transferability_critic_framework} and facilitate the category-wise alignment in Section \ref{sec:soft_divergence_loss}, we employ a confidence-guided pseudo labels generator \cite{Zou_2019_ICCV, Zou_2018_ECCV} to produce soft pseudo labels for unlabeled target data. It efficiently attenuates the enormous deviation brought by unconfident or invalid supervision of pseudo labels. A feasible solution is to progressively mine valid labels relying on their confidence score along the training process. Specifically, after the source sample $x_i^s$ with label $y_i^s$ are utilized to train $E$ and $C$, the probability output of target sample $x_j^t$ via $C$ is regarded as confidence score. The generation of valid soft pseudo labels has a preference for pixels with high confidence scores. Generally, the joint objective function for training with source data and pseudo labels selection is formulated as follows: 
\begin{equation}
\begin{split}
&\min_{\theta_{E}, \theta_{C}}\mathcal{L}_{seg} = \mathbb{E}_{(x_i^s, y_i^s)\in D_s}\big[\sum\limits_{i=1}^{n_s} \sum\limits_{m=1}^{|x_i^s|} \sum\limits_{k=1}^{K}\big(- (y_i^s)_m^k \mr{log}(C(x_i^s; \theta_C)_m^k) \big) \big] +  \\
&\quad~~  \mathbb{E}_{(x_j^t, \hat{y}_j^t)\in D_t}\big[\sum\limits_{j=1}^{n_t} \sum\limits_{n=1}^{|x_j^t|} \sum\limits_{k=1}^{K} \big( - (\hat{y}_j^t)_n^k \mr{log}(\frac{C(x_j^t; \theta_C)_n^k}{\delta_k}) + \gamma(\hat{y}_j^t)_n^k \mr{log}((\hat{y}_j^t)_n^k) \big) \big], \\
&\quad~~ s.t. ~(\hat{y}_j^t)_n = [(\hat{y}_j^t)_n^1, \dots, (\hat{y}_j^t)_n^K] \in \big\{\{\mathbb{C}^K|\sum\limits_{k=1}^K (\hat{y}_j^t)_n^k =1, \mathbb{C}^K\in\mathbb{R}^{K} \} \cup \mathbf{0}\big\},
\end{split}
\label{eq:pseudo_labels_objective}
\end{equation}
where the first term is supervised training for source data $D_s$, and the second term is used for pseudo labels selection of target data $D_t$. $\gamma \geq 0$ is a balanced parameter. $\theta_C$ denotes the parameters of classifier $C$. $\mathbb{C}^K$ represents the $K$ dimensional vector in continuous probability space. $K$ is the number of classes. $y_i^s$ and $\hat{y}_j^t$ denote the one-hot encoding groundtruth of $x_i^s$ and soft pseudo label of $x_j^t$. Note that the values of $\hat{y}_j^t$ are in continuous space rather than discrete space (\emph{i.e.}, one-hot labels). $(y_i^s)_m^k$ and $(\hat{y}_j^t)_n^k$ denote the values of the $k$-th category in the $m$-th pixel of $x_i^s$ and the $n$-th pixel of $x_j^t$, respectively. Likewise, $C(x_i^s; \theta_C)_m^k$ and $C(x_j^t; \theta_C)_n^k$ are the probabilities predicted as the $k$-th class via $C$ for the $m$-th pixel of $x_i^s$ and the $n$-th pixel of $x_j^t$. $\delta_k (k=1, \dots, K)$ are the selection thresholds of pseudo labels for each class, which are adaptively determined in \textbf{Algorithm 1}. The last part in the second item of Eq.~\eqref{eq:pseudo_labels_objective} encourages the continuity of output probabilities, and prevents the misleading supervision of invalid or unconfident pseudo labels.

Intuitively, $\delta_k$ determines the selection amount of confident pseudo labels belonging to the $k$-th class, and the larger value of $\delta_k$ facilitates the selection of more confident pseudo labels along the training process. More importantly, as shown in Eq.~\eqref{eq:pseudo_labels_objective}, it forcefully regards unconfident pseudo labels as invalid labels (\emph{i.e.}, $(\hat{y}_j^t)_n = \mathbf{0}$) while preventing the trivial solution from assigning all pseudo pixel labels as $\mathbf{0}$. Note that allocating $(\hat{y}_j^t)_n$ as $\mathbf{0}$ can neglect this unconfident pseudo label of the $n$-th pixel in the training phase. The determination of $\delta_k$ is summarized in \textbf{Algorithm 1}, where $\mathcal{S_A}$ is initialized as 35\% and increased by 5\% in each epoch until the maximum value 50\%.

\renewcommand{\algorithmicrequire}{\textbf{Input:}}
\renewcommand{\algorithmicensure}{\textbf{Output:}}
\begin{algorithm}[t]			
	\caption{The Determination of $\delta_k$ in Eq.~\eqref{eq:pseudo_labels_objective}}
	\begin{algorithmic}[1]
		\REQUIRE The number of classes $K$, selection amount $\mathcal{S_A}$ of soft pseudo labels, classifier $C$, target dataset $D_t = \{x_j^t\}_{j=1}^{n_t}$; 
		\ENSURE $\delta_k ~(k=1, \dots, K)$    \\
		\FOR {$j=1,\ldots,n_t$}
		\STATE  $MP_j = C(x_j^t, \theta_{C})$; 
		\qquad  \# Maximum probability output of each pixel;		
		\STATE  $Y_j = \mr{argmax}(C(x_j^t, \theta_{C}), \mr{axis}=3)$; 
		\qquad \# Category prediction of each pixel;
		\FOR {$k=1,\ldots,K$}
		\STATE  $L_k = [L_k, \mr{matrix\_to\_vector}(MP_j(Y_j == k))]$;
		\# Probabilities predicted as $k$; 
		\ENDFOR  \\
		\ENDFOR \\
		\FOR {$k=1,\ldots,K$}
		\STATE  $S_k = \mr{sorting}(L_k, \mr{descending})$;
		\qquad \# Sort in a descending order
		\STATE  $\delta_k = S_k[\mathcal{S_A}\cdot\mr{length}(S_k)]$; 
		~\# Select probability ranked at $\mathcal{S_A}\cdot\mr{length}(S_k)$ as $\delta_k$;
		\ENDFOR \\		
		return $\delta_k~(k=1, \dots, K)$. 
	\end{algorithmic}
\end{algorithm}

To progressively mine confident soft pseudo labels, the first step is to optimize the second term of Eq.~\eqref{eq:pseudo_labels_objective} by employing a Lagrange multiplier $\psi$. For simplification, the formulation in term of the $n$-th pixel of $x_j^t$ is rewritten as: 
\begin{equation}
\begin{split}
\min_{(\hat{y}_j^t)_n^k} \sum\limits_{k=1}^{K} \big( - (\hat{y}_j^t)_n^k \mr{log}(\frac{C(x_j^t; \theta_C)_n^k}{\delta_k}) + \gamma(\hat{y}_j^t)_n^k \mr{log}((\hat{y}_j^t)_n^k) \big) + \psi \big( \sum\limits_{k=1}^K (\hat{y}_j^t)_n^k -1 \big),
\end{split}
\label{eq:pseudo_labels_optimization} 
\end{equation}
where the first item of Eq.~\eqref{eq:pseudo_labels_optimization} is defined as selection cost $\mathcal{S_C}((\hat{y}_j^t)_n)$. The optimal solution of $(\hat{y}_j^t)_n^k$ can be relaxedly achieved by setting the gradient equal to 0 with respect to the $k$-th category, \emph{i.e.},
\begin{equation}
\begin{split}
(\hat{y}_j^t)_n^k = e^{-\frac{\gamma + \psi}{\gamma}} \big( \frac{C(x_j^t; \theta_C)_n^k}{\delta_k} \big)^{\frac{1}{\gamma}}, ~\forall k=1, \dots,K,
\end{split}
\label{eq:pseudo_labels_solution} 
\end{equation}
where $e^{-\frac{\gamma + \psi}{\gamma}} = 1 \big/ \big[ \sum_{k=1}^K \big( \frac{C(x_j^t; \theta_C)_n^k}{\delta_k} \big)^{\frac{1}{\gamma}}\big]$ is determined via the Eq.~\eqref{eq:pseudo_labels_solution} and the constrain $\sum_{k=1}^K (\hat{y}_j^t)_n^k =1$. The second step involves selecting $(\hat{y}_j^t)_n$ or $\mathbf{0}$ as pseudo label by judging which one leads to a lower selection cost $\mathcal{S_C}$. Thus, the final solution for selecting pseudo labels is:
\begin{equation}
\begin{split}
(\hat{y}_{j}^{t})_n^k =
\left\{
\begin{aligned}		
&\big(\frac{C(x_j^t; \theta_C)_n^k}{\delta_k} \big)^{\frac{1}{\gamma}} \Big/ \big[ \sum\limits_{k=1}^K \big( \frac{C(x_j^t; \theta_C)_n^k}{\delta_k} \big)^{\frac{1}{\gamma}}\big], ~~ \mr{if} ~ \mathcal{S_C}((\hat{y}_j^t)_n) ~\textless~ \mathcal{S_C}(\mathbf{0}),  \\
&0, \qquad\qquad\qquad \mr{otherwise}.  \\
\end{aligned} 					
\right.											
\end{split}							
\label{equation:pseudo_labels_final_solution}	
\end{equation}

\subsection{Symmetric Soft Divergence Loss $\mathcal{L}_{div}$} 
\label{sec:soft_divergence_loss}
With the confident pseudo labels mined by Eq.~\eqref{equation:pseudo_labels_final_solution} in Section \ref{sec:pseudo_labels_generator} as reliable guidance, a symmetric soft divergence loss $\mathcal{L}_{div}$ is proposed to align category-wise distribution. In other words, $\mathcal{L}_{div}$ encourages semantic features of same class to be compactly clustered together via category labels regardless of domains, and drives features from different classes across domains to satisfy the inter-category relationships. Different from \cite{NIPS2017_6963, 10.5555/3042817.3043028} that align the feature centroids directly, $\mathcal{L}_{div}$ aims to mitigate the conditional distributions shift by minimizing the symmetric soft Kullback-Leibler (KL) divergence among source and target features, with respect to each class $k$. Furthermore, $\mathcal{L}_{div}$ enforces category ambiguities across domains to be more consistent, which is concretely written as: 
\begin{equation}
\begin{split}
\min_{\theta_{E}}\mathcal{L}_{div} = \frac{1}{K} \sum\limits_{k=1}^K \frac{1}{2}\big(D_{\mr{KL}}(F_s^k||F_t^k) + D_{\mr{KL}}(F_t^k||F_s^k)\big);
\end{split}
\label{eq:soft_divergence_loss} 
\end{equation}
where $D_{\mr{KL}}(F_s^k||F_t^k) = \sum_q (F_s^k)_q \mr{log}\frac{(F_s^k)_q}{(F_t^k)_q}$ denotes the KL divergence between source feature centroid $F_s^k$ and target soft feature centroid $F_t^k$, in terms of the $k$-th category. Likewise, $D_{\mr{KL}}(F_t^k||F_s^k)=\sum_q (F_t^k)_q \mr{log}\frac{(F_t^k)_q}{(F_s^k)_q}$ shares the similar definition with $D_{\mr{KL}}(F_s^k||F_t^k)$. According to the source label $y_i^s$, the source feature centroid $F_s^k$ of the $k$-th class is computed by the following equation, \emph{i.e.},
\begin{equation}
\begin{split}
F_s^k = \mathbb{E}_{(x_i^s, y_i^s)\in D_s}\big[\frac{1}{N_s^k} \sum\limits_{i=1}^{n_s} \sum\limits_{m=1}^{|x_i^s|} \big( E(x_i^s; \theta_{E})_m \cdot \mathbf{1}_{\mr{argmax}((y_i^s)_m) = k} \big) \big] ;
\end{split}
\label{eq:feature_centroid_target} 
\end{equation}
where $N_s^k= \sum_{i=1}^{n_s} \sum_{m=1}^{|x_i^s|} \mathbf{1}_{\mr{argmax}((y_i^s)_m) = k}$ is the number of pixels annotated as class $k$ in dataset $D_s$. $E(x_i^s; \theta_{E})_m$ represents the feature extracted by $E$ at the $m$-th pixel in source sample $x_i^s$. Similarly, with the assistance of generated pseudo labels $\hat{y}_j^t$, the target soft feature centroid $F_t^k$ of the $k$-th category is defined as:
\begin{equation}
\begin{split}
F_t^k = \mathbb{E}_{(x_j^t, \hat{y}_j^t)\in D_t}\big[\frac{1}{N_t^k} \sum\limits_{j=1}^{n_t} \sum\limits_{n=1}^{|x_j^t|} (\hat{y}_j^t)_n^k E(x_j^t; \theta_{E})_n \cdot \mathbf{1}_{\mr{argmax}((\hat{y}_j^t)_n)=k} \big],
\end{split}
\label{eq:feature_centroid_source} 
\end{equation}
where $N_t^k = \sum_{j=1}^{n_t} \sum_{n=1}^{|x_j^t|} (\hat{y}_j^t)_n^k \cdot \mathbf{1}_{\mr{argmax}((\hat{y}_j^t)_n) = k}$ is the sum of the probabilities predicted as class $k$ in dataset $D_t$. $E(x_j^t; \theta_{E})_n$ denotes the  feature extracted by $E$ at the $n$-th pixel in target image $x_j^t$.

\subsection{Implementation Details} 
\textbf{Network Architecture:} 
The DeepLab-v2 \cite{net:deeplab} with ResNet-101 \cite{net:resnet} and FCN-8s \cite{Shelhamer:2017:FCN:3069214.3069246} with VGG-16 \cite{Simonyan15} are employed as our backbone network, \emph{i.e.}, $E$ and $C$. They are initially pre-trained by the ImageNet \cite{5206848}. The discriminator $D$ contains 5 convolutional layers with channels as \{64, 128, 256, 512, 1\}, where each block excluding the last one is activated by the leaky ReLU with parameter as 0.2. As for $T_C$, the state branch includes 3 layers whose channels are \{64, 32, 16\}, and the channels of two convolutional blocks in policy branch are both 16. Moreover, there is only one convolution layer with the channel as 1 in $T_Q$.

\renewcommand{\algorithmicrequire}{\textbf{Input:}}
\renewcommand{\algorithmicensure}{\textbf{Output:}}
\begin{algorithm}[t]			
	\caption{The Optimization Procedure of Our CSCL Model}
	\begin{algorithmic}[1]
		\REQUIRE The source and target data ($D_s$ and $D_t$), maximum iteration $I$,  $\gamma, \xi_1, \xi_2$, $\xi_3$; 
		\ENSURE The network parameters $\theta_{E}, \theta_{C}, \theta_{D}, \theta_{T_Q}$ and $\theta_{T_C}$  \\
		\STATE  Initialize all parameters of network architecture;
		\FOR {$i=1,\ldots,I$}
		\STATE  Randomly select a batch of samples from both $D_s$ and $D_t$; 
		\STATE  Update $\theta_E$ and $\theta_{C}$ via minimizing the first term of $\mathcal{L}_{seg}$;
		\STATE  Generate confident soft pseudo labels for target samples via Eq.~\eqref{equation:pseudo_labels_final_solution}; 
		\STATE  Update $\theta_E$ and $\theta_{D}$ via optimizing $\min_{\theta_{E}}\max_{\theta_D}~\xi_2\mathcal{L}_{adv} +\xi_3\mathcal{L}_{div}$;
		\STATE  Update $\theta_{T_Q}$ via minimizing $\mathcal{L}_{seg} + \xi_1\mathcal{L}_{cri}$;
		\STATE  Update $\theta_{T_C}$ via minimizing $\mathcal{L}_{reg}$;
		\ENDFOR   \\
		Return $\theta_{E}, \theta_{C}, \theta_{D}, \theta_{T_Q}$ and $\theta_{T_C}$. 
	\end{algorithmic} 
\end{algorithm}

\textbf{Training and Evaluating:} 
In the training phase, the overall optimization objective $\mathcal{L}_{obj}$ for our proposed CSCL model is formulated as: 
\begin{equation}
\begin{split}
\min_{\theta_{E}, \theta_{C}, \theta_{T_Q}, \theta_{T_C}} \max_{\theta_{D}}~ \mathcal{L}_{obj} = \mathcal{L}_{seg} + \mathcal{L}_{reg}+ \xi_1\mathcal{L}_{cri} + \xi_2 \mathcal{L}_{adv} + \xi_3\mathcal{L}_{div}, 
\end{split}
\label{eq:overall_optimization_objective} 
\end{equation}
where $\xi_1, \xi_2$, $\xi_3 \geq 0$ are balanced weights. The optimization procedure is summarized in \textbf{Algorithm 2}. For DeepLab-v2 with ResNet-101, SGD is utilized as optimizer whose the initial learning rate is $2.5 \times 10^{-4}$ and decreased via poly policy with power as 0.9. For FCN-8s with VGG-16, we employ Adam as optimizer whose initial learning rate is $1.0\times 10^{-4}$ and the momentum is set as 0.9 and 0.99. Moreover, Adam is also used to optimize the discriminator $D$. The initial learning rate is respectively set as $1.5\times 10^{-4}$ and $1.5\times 10^{-6}$ for ResNet-101 and VGG-16. We empirically set $\gamma=0.25$ in Eq.~\eqref{equation:pseudo_labels_final_solution}, and set $\xi_1=0.3, \xi_2=0.001$ and $\xi_3=10$ in Eq.~\eqref{eq:overall_optimization_objective} for all experiments. The batch size for training is set as 1. In the evaluating stage, we directly forward $x_j^t$ into $E$ and $C$ for prediction.

\section{Experiments} 
\subsection{Datasets and Evaluation Metric}	
\textbf{Cityscapes} \cite{Cordts_2016_CVPR} is a real-world dataset about street scenes from 50 different European cities, which is divided into a training subset with 2993 samples, a testing subset with 1531 images and a validation subset with 503 samples. There are total 34 distinct finely-annotated categories in this dataset.

\textbf{GTA} \cite{Richter_2016_ECCV} with 24996 images is generated from a fictional computer game called Grand Theft Auto V, whose 19 classes are compatible with Cityscapes \cite{Cordts_2016_CVPR}.  

\textbf{SYNTHIA} \cite{Ros_2016_CVPR} is a large-scale automatically-labeled synthetic dataset for semantic segmentation task of urban scenes, whose the subset named SYNTHIA-RAND-CITYSCAPES with 9400 images is used in our experiments.

\textbf{NTHU} \cite{exp:CCA} consists of four real world datasets, which are respectively collected from Rome, Rio, Tokyo and Taipei. Every dataset contains the training and testing subsets, and shares 13 common categories with Cityscapes \cite{Cordts_2016_CVPR}. 

\textbf{Evaluation Metric:} Intersection over Union (IoU) is employed as evaluation metric, and mIoU denotes the mean value of IoU.

\begin{table*}[t]
	\centering
	\setlength{\tabcolsep}{1.175mm}	
	\caption{Adaptation performance of transferring from GTA \cite{Richter_2016_ECCV} to Cityscapes \cite{Cordts_2016_CVPR}.}
	\scalebox{0.520}{
		\begin{tabular}{|c|c|ccccccccccccccccccc|c|}
			\hline
			Method & Net & road & sidewalk & building & wall & fence & pole & light & sign & veg & terrain & sky & person & rider & car & truck & bus & train & mbike & bike & mIoU  \\	
			\hline
			\hline
			
			Source only \cite{Simonyan15} &  & 18.1 & 6.8 & 64.1 & 7.3 & 8.7 & 21.0 & 14.9 & 16.8 & 45.9 & 2.4 & 64.4 & 41.6 & 17.5 & 55.3 & 8.4 & 5.0 & 6.9 & 4.3 & 13.8 & 22.3 \\ 
			
			Wild \cite{exp:Wild} & & 70.4 & 32.4 & 62.1 & 14.9 & 5.4 & 10.9 & 14.2 & 2.7 & 79.2 & 21.3 & 64.6 & 44.1 & 4.2 & 70.4 & 8.0 & 7.3& 0.0 & 3.5 & 0.0 & 27.1 \\ 
			
			CDA \cite{exp:CL} &  & 74.9 & 22.0 & 71.7 & 6.0 & 11.9 & 8.4 & 16.3 & 11.1 & 75.7 & 11.3 & 66.5 & 38.0 & 9.3 & 55.2 & 18.8 & 18.9 & 0.0 & 16.8 & 14.6 & 28.9  \\

			MCD \cite{Saito_2018_CVPR} & & 86.4 & 8.5 & 76.1 & 18.6 & 9.7 & 14.9 & 7.8 & 0.6 & 82.8 & 32.7 & 71.4 & 25.2 & 1.1 & 76.3 & 16.1 & 17.1 & 1.4 & 0.2 & 0.0 & 28.8  \\			
			
			CBST \cite{Zou_2018_ECCV} & & 66.7 & 26.8 & 73.7 & 14.8 & 9.5 & 28.3 & 25.9 & 10.1 & 75.5 & 15.7 & 51.6 & 47.2 & 6.2 & 71.9 & 3.7 & 2.2 & 5.4 & 18.9 & \textbf{32.4} & 30.9 \\
			
			CLAN \cite{Luo_2019_CVPR} & & 88.0 & 30.6& 79.2 &23.4 &20.5 &26.1& 23.0 &14.8 &81.6 &34.5 &72.0 &45.8& 7.9& 80.5& 26.6 &29.9& 0.0& 10.7& 0.0 &36.6 \\
			
			SWD \cite{Lee_2019_CVPR} &  & \textbf{91.0}& 35.7& 78.0 &21.6 &21.7 &\textbf{31.8} &\textbf{30.2} & \textbf{25.2}& 80.2& 23.9& 74.1& 53.1& 15.8& 79.3& 22.1 &26.5& 1.5 &17.2& 30.4& 39.9 \\
			
			ADV \cite{Vu_2019_CVPR} & & 86.9 &28.7 &78.7 &28.5& 25.2 &17.1& 20.3& 10.9& 80.0& 26.4& 70.2& 47.1& 8.4& 81.5& 26.0& 17.2& \textbf{18.9} & 11.7& 1.6 &36.1 \\
			
			DPR \cite{Tsai_2019_ICCV} & &87.3& 35.7 &79.5 &\textbf{32.0} & 14.5& 21.5& 24.8& 13.7& 80.4 &32.0 &70.5 &50.5& 16.9& 81.0 &20.8 &28.1& 4.1& 15.5& 4.1& 37.5 \\
			
			LDF \cite{Lee_2019_ICCV} & & 88.8& 36.9& 76.9& 20.9& 15.4 &19.6& 21.8& 7.9& 82.9& 26.7& 76.1& 51.7& 9.4& 76.1& 22.4 &28.9& 1.7& 15.2& 0.0& 35.8 \\
			
			SSF \cite{Du_2019_ICCV} & & 88.7& 32.1& 79.5& 29.9 &22.0& 23.8& 21.7& 10.7& 80.8& 29.8 &72.5& 49.5& 16.1& \textbf{82.1}& 23.2& 18.1& 3.5& 24.4 &8.1& 37.7 \\
			
			PyCDA \cite{Lian_2019_ICCV} & & 86.7 &24.8& \textbf{80.9}& 21.4& 27.3& 30.2& 26.6& 21.1& \textbf{86.6}& 28.9& 58.8& \textbf{53.2}& 17.9& 80.4& 18.8& 22.4& 4.1& 9.7& 6.2& 37.2 \\

			\rowcolor{lightgray}
			Ours-w/oTC &  & 88.7 & 43.5 & 74.2 & 29.4 & 26.1 & 10.8 & 17.9 & 16.0 & 78.9 & 38.7 & 75.2 & 38.4 & 17.5 & 73.3 & 29.5 & 37.8 & 0.4 & 26.7 & 29.3 & 39.6 \\ 
			
			\rowcolor{lightgray}
			Ours-w/oCG &  & 88.3 & 39.1 & 76.4 & 28.3 & 22.4 & 11.9 & 17.3 & 18.7 & 76.2 & 36.0 & 74.5 & 39.3 & 18.6 & 73.0 & \textbf{30.7} & \textbf{36.1} & 0.8 & 27.4 & 30.1 & 39.2 \\
			
			\rowcolor{lightgray}
			Ours-w/oSD & & 89.1 & 45.0 & 77.6 & 28.8 & 27.5 & 12.1 & 18.6 & 21.2 & 77.4 & 37.7 & 76.5 & 40.7 & \textbf{19.8} & 74.4 & 29.3 & 33.8 & 0.6 & 28.2 & 29.2 & 40.4  \\
			
			\rowcolor{lightgray}
			Ours & & 89.8 & \textbf{46.1} & 75.2 & 30.1 & \textbf{27.9} & 15.0 & 20.4 & 18.9 & 82.6 & \textbf{39.1} & \textbf{77.6} & 47.8 & 17.4 & 76.2 & 28.5 & 33.4 &0.5 & \textbf{29.4} & 30.8& \textbf{41.4} \\
			
			\hline
			\hline
			
			Source only \cite{net:resnet} &  & 75.8 & 16.8 & 77.2 & 12.5 & 21.0 & 25.5 & 30.1 & 20.1 & 81.3 & 24.6 & 70.3 & 53.8 & 26.4 & 49.9 & 17.2 & 25.9 & 6.5 & 25.3 & 36.0 & 36.6  \\  
			
			LtA \cite{exp:LtA} &  & 86.5 & 36.0 & 79.9 & 23.4 & 23.3 &23.9 &35.2 &14.8 & 83.4& 33.3 & 75.6 & 58.5 & 27.6 & 73.7 & 32.5 & 35.4 & 3.9 & 30.1 & 28.1 & 42.4 \\
			
			CGAN \cite{exp:CGAN} & &89.2 & 49.0 & 70.7&13.5 & 10.9 & 38.5 & 29.4 & 33.7& 77.9 & 37.6 & 65.8 & \textbf{75.1} & 32.4& 77.8 & \textbf{39.2}&45.2 & 0.0& 25.2 & 35.4 & 44.5  \\
			
			CBST \cite{Zou_2018_ECCV} &  & 88.0 & \textbf{56.2} & 77.0 & 27.4 & 22.4 & \textbf{40.7} & \textbf{47.3} & 40.9 & 82.4 & 21.6 & 60.3 & 50.2 & 20.4 & 83.8 & 35.0 & \textbf{51.0} & 15.2 & 20.6 & 37.0 & 46.2 \\
			
			DLOW \cite{Gong_2019_CVPR} & & 87.1 & 33.5 & 80.5 & 24.5 & 13.2 & 29.8 & 29.5 & 26.6 & 82.6 & 26.7 & 81.8& 55.9 & 25.3 & 78.0 & 33.5 & 38.7 & 0.0 & 22.9 & 34.5 & 42.3  \\
			
			CLAN \cite{Luo_2019_CVPR} & &87.0 & 27.1 & 79.6 & 27.3 & 23.3 & 28.3 & 35.5 & 24.2 & 83.6 & 27.4 & 74.2 & 58.6 & 28.0 & 76.2 & 33.1 & 36.7 & 6.7 & 31.9 & 31.4 & 43.2  \\
			
			SWD \cite{Lee_2019_CVPR} & & 92.0 & 46.4 & 82.4 & 24.8 & 24.0 & 35.1 & 33.4 & 34.2 & 83.6 & 30.4 & 80.9 & 56.9 & 21.9 & 82.0 & 24.4 & 28.7 & 6.1 & 25.0 & 33.6 & 44.5 \\
			
			ADV \cite{Vu_2019_CVPR} & & 89.4 & 33.1 & 81.0 & 26.6 & 26.8 & 27.2 & 33.5 & 24.7 & 83.9 & 36.7 & 78.8 & 58.7 & 30.5 & 84.8 & 38.5 & 44.5 & 1.7 & 31.6 & 32.5 & 45.5 \\
			
			SWLS \cite{Dong_2019_ICCV} & ~~~\begin{rotate}{90} ResNet
			\end{rotate} & \textbf{92.7} &48.0 &78.8 & 25.7 & 27.2 & 36.0 & 42.2 & \textbf{45.3} & 80.6 &14.6 &66.0 & 62.1& 30.4 &\textbf{86.2} & 28.0 &45.6 &\textbf{35.9} &16.8 & 34.7 & 47.2 \\
			
			DPR \cite{Tsai_2019_ICCV} & & 92.3 & 51.9 & 82.1 & 29.2 & 25.1 & 24.5 & 33.8 & 33.0 & 82.4 & 32.8 & 82.2 & 58.6 & 27.2 & 84.3 & 33.4 & 46.3 & 2.2 & 29.5 & 32.3 & 46.5  \\ 
			
			SSF \cite{Du_2019_ICCV} & & 90.3& 38.9& 81.7& 24.8& 22.9 &30.5& 37.0& 21.2& 84.8& 38.8& 76.9& 58.8& 30.7& 85.7& 30.6& 38.1& 5.9& 28.3& 36.9& 45.4  \\
			
			PyCDA \cite{Lian_2019_ICCV} & & 90.5 & 36.3 & 84.4 & 32.4 & \textbf{28.7} & 34.6 & 36.4 & 31.5 & \textbf{86.8} & 37.9 & 78.5 & 62.3 & 21.5 & 85.6 & 27.9 & 34.8 & 18.0 & 22.9 & \textbf{49.3} & 47.4 \\ 
			
			\rowcolor{lightgray}
			Ours-w/oTC &  & 91.6 & 47.9 & 83.4 & 35.0 & 23.6 & 31.6 & 36.5 & 31.9 & 82.9 & 36.6 & 76.4 & 58.7 & 25.6 & 81.5 & 37.1 & 46.6 & 0.5 & 26.0 & 34.0 & 46.7 \\
			
			\rowcolor{lightgray}
			Ours-w/oCG & & 89.6& 40.8 & \textbf{84.6} & 30.4 & 22.7 &32.0 & 37.4 & 33.7 &82.3 & 39.6 & 80.7 & 57.4 & 28.7 & 82.8 & 27.4 & 48.2 & 1.0 & 27.0 & 29.5 & 46.1  \\
			
			\rowcolor{lightgray}
			Ours-w/oSD & & 89.3 & 47.8 & 82.4 & 31.3 & 25.1 & 31.2 & 37.3 & 34.9 & 83.9 & 37.9 & 83.0 & 59.4 & 31.4 & 79.0 & 35.7 & 42.0 & 0.2 & 34.1 & 34.6 & 47.4 \\
			
			\rowcolor{lightgray}
			Ours & & 89.6 & 50.4 & 83.0 & \textbf{35.6} & 26.9 & 31.1 & 37.3 & 35.1 & 83.5 & \textbf{40.6} & \textbf{84.0} & 60.6 & \textbf{34.3} & 80.9 & 35.1 & 47.3 & 0.5 & \textbf{34.5} & 33.7 & \textbf{48.6}  \\
			\hline
			
		\end{tabular}
	}			
	\label{tab:Exp_GTA2City} 
\end{table*}

\subsection{Experiments on GTA $\rightarrow$ Cityscapes Task}
When transferring from  GTA \cite{Richter_2016_ECCV} to Cityscapes \cite{Cordts_2016_CVPR}, GTA and the training subset of Cityscapes are respectively considered as source and target domains. The validation subset of Cityscapes is used for evaluation.

\textbf{Comparisons Performance:} Table~\ref{tab:Exp_GTA2City} reports the adaptation performance of our model compared with state-of-the-art methods on GTA $\rightarrow$ Cityscapes task. From the Table~\ref{tab:Exp_GTA2City}, we have the following observations: 1) Our model outperforms all existing state-of-the-art methods about 1.2\% $\sim$ 14.3\% mIoU, which efficiently matches both domain-wise and category-wise semantic consistency across domains. 2) For the classes with various appearances among different datasets (\emph{e.g.}, rider, motorbike, terrain, fence and sidewalk), our model achieves better performance to mitigate the large distribution shifts by exploring transferable representations.

\textbf{Ablation Studies:} 
In this subsection, we investigate the importance of different components in our model by conducting variant experiments, as shown in the gray part of Table~\ref{tab:Exp_GTA2City}. Training the model without critical transfer based adversarial framework, confidence-guided pseudo labels generator and symmetric soft divergence loss are denoted as Ours-w/oTC, Ours-w/oCG and Ours-w/oSD, respectively. The transfer performance degrades 1.0\% $\sim$ 2.5\% when any component of our model is removed. Table~\ref{tab:Exp_GTA2City} validates that all designed components play an importance role in exploring transferable representations while neglecting irrelevant knowledge. Moreover, they can well narrow both marginal and conditional distributions shifts across domains.

\begin{figure*}[t]
	\begin{minipage}[t]{0.24\linewidth}
		\centering
		\includegraphics[trim = 54mm 76mm 74mm 154mm, clip, height=78pt, width=92pt]{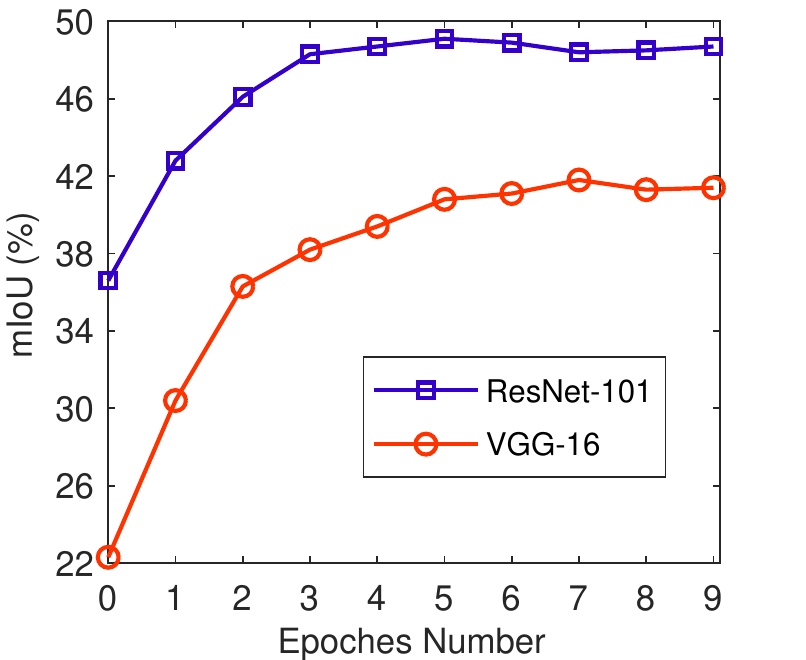}
		\\ ~ (a) mIoU
		\label{fig:GTA_mIoU}
	\end{minipage}
	\begin{minipage}[t]{0.24\linewidth}
		\centering
		\includegraphics[trim = 54mm 76mm 74mm 154mm, clip, height=78pt, width=92pt]{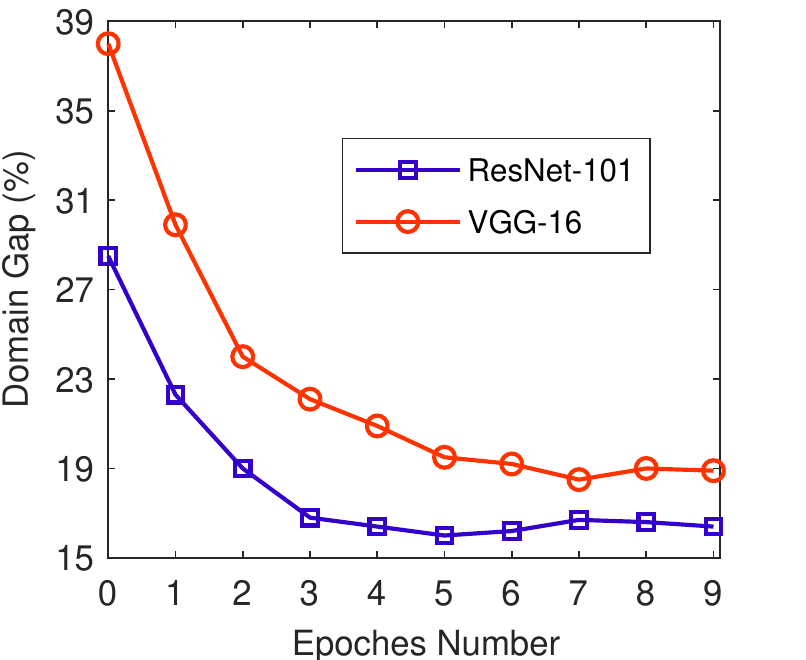}
		\\ ~ (b) Domain gap
		\label{fig:GTA_domain_gap}
	\end{minipage}
	\begin{minipage}[t]{0.24\linewidth}
		\centering
		\includegraphics[trim = 54mm 80mm 73mm 154mm, clip, height=80pt, width=95pt]{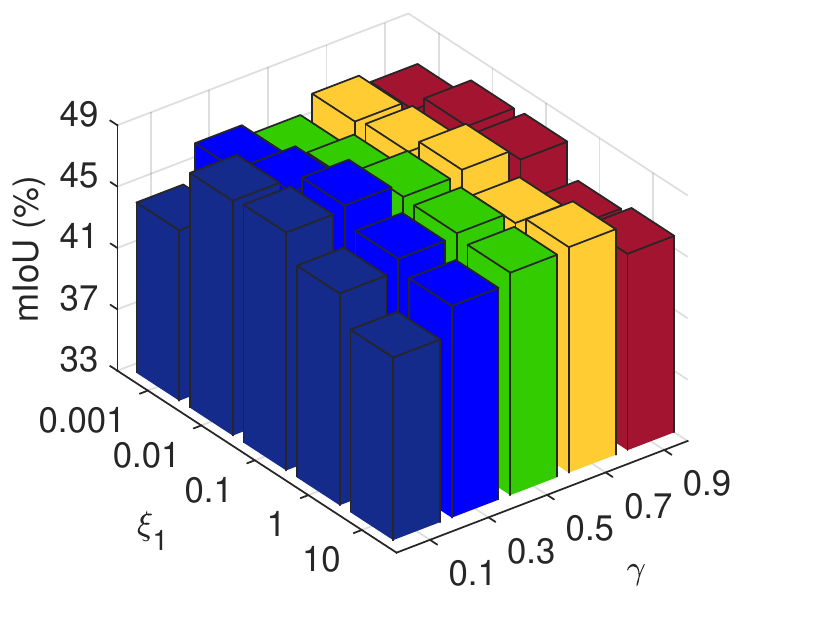}
		\\ (c) Effect of $\{\xi_1, \gamma\}$
		\label{fig:GTA_parameter_xi1_gamma}
	\end{minipage}
	\begin{minipage}[t]{0.24\linewidth}
		\centering
		\includegraphics[trim = 54mm 80mm 73mm 154mm, clip, height=80pt, width=95pt]{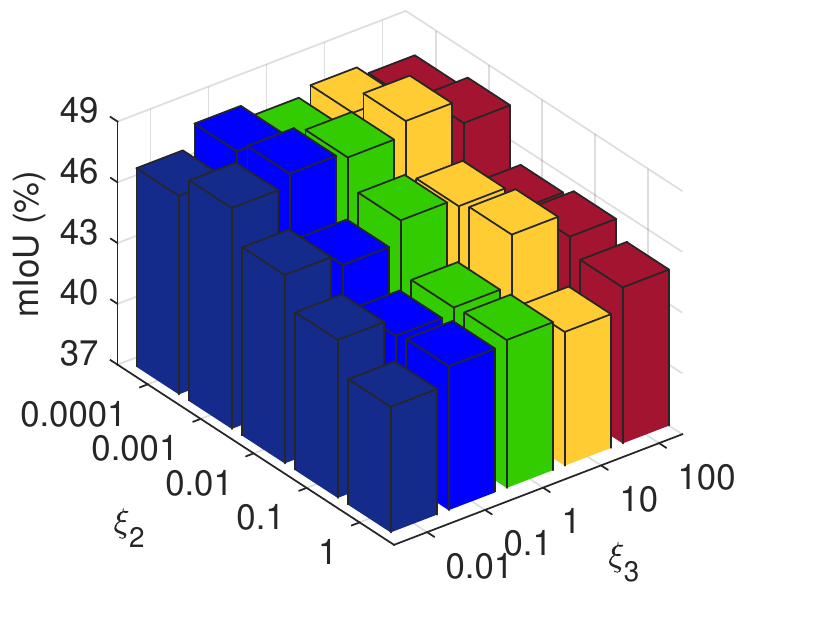}
		\\ (d) Effect of $\{\xi_2, \xi_3\}$
		\label{fig:GTA_parameter_xi2_xi3}
	\end{minipage}
	\caption{Experimental results on GTA $\rightarrow$ Cityscapes task. (a): 
		Convergence curves about mIoU; (b): Convergence curves about domain gap; (c): The effect of $\{\xi_1, \gamma\}$ when $\xi_2=0.001$ and $\xi_3=10$; (d): The effect of $\{\xi_2, \xi_3\}$ when $\xi_1=0.3$ and $\gamma=0.25$.}
	\label{fig:convergence_and_parameter} 
\end{figure*}

\textbf{Convergence Investigation:} 
The convergence curves of our model with respect to mIoU and domain gap are respectively depicted in Fig.~\ref{fig:convergence_and_parameter} (a) and Fig.~\ref{fig:convergence_and_parameter} (b). Specifically, our model equipped with VGG-16 or ResNet-101 can achieve efficient convergence when the iterative epoch number is 5. More importantly, domain gap among different datasets is iteratively minimized to a stable value via our model, which efficiently narrows the domain discrepancy.

\textbf{Parameter Sensitivity:} 
Extensive parameter experiments are empirically conducted to investigate the parameter sensitivity, and determine the optimal selection of hyper-parameters. Fig.~\ref{fig:convergence_and_parameter} (c) and Fig.~\ref{fig:convergence_and_parameter} (d) present the effects of hyper-parameters $\{\xi_1, \gamma\}$ and $\{\xi_2, \xi_3\}$ on our model with ResNet-101 as basic network, respectively. We can conclude that our model achieves stable transfer performance even though hyper-parameters have a wide range of selection. Moreover, $\gamma$ is essential to mine confident pseudo labels for target samples, which provides positive guidance for transferability-critic $T_C$ and $\mathcal{L}_{div}$.

\begin{table*}[t]	
	\centering
	\setlength{\tabcolsep}{1.33mm}
	\caption{Adaptation performance of transferring from SYNTHIA \cite{Ros_2016_CVPR} to Cityscapes \cite{Cordts_2016_CVPR}.}
	\scalebox{0.575}{
		\begin{tabular}{|c|c|cccccccccccccccc|c|}
			\hline
			Method & Net & road & sidewalk & building & wall & fence & pole &light &sign & veg & sky & person & rider & car & bus & mbike & bike & mIoU \\
			\hline			
			\hline	
			
			Source only \cite{Simonyan15} &  & 6.4 & 17.7 & 29.7 & 1.2 & 0.0 & 15.1 & 0.0 & 7.2 & 30.3 & 66.8 & 51.1 & 1.5 & 47.3 & 3.9 & 0.1 & 0.0 &17.4 \\
			
			Wild \cite{exp:Wild} &  & 11.5 & 19.6 & 30.8 & 4.4 & 0.0 & 20.3 & 0.1 & 11.7& 42.3 & 68.7 & 51.2 & 3.8 & 54.0 & 3.2 & 0.2 & 0.6 & 20.2  \\
			
			CDA \cite{exp:CL} & & 65.2 & 26.1 & 74.9 & 0.1 & 0.5 & 10.7& 3.7 & 3.0 &76.1 & 70.6 & 47.1 & 8.2 & 43.2 & 20.7 & 0.7 & 13.1 & 29.0 \\ 	
			
			LSD \cite{exp:LSD} &  & 80.1 &29.1 &77.5 &2.8 &0.4 &26.8 &11.1 &18.0 &78.1 &76.7 &48.2 & 15.2&70.5 &17.4 &8.7 &16.7 &36.1 \\
			
			DCAN \cite{Wu_2018_ECCV} & & 79.9 &30.4& 70.8& 1.6& 0.6& 22.3& 6.7& \textbf{23.0} &76.9& 73.9& 41.9 &16.7& 61.7& 11.5& 10.3& \textbf{38.6}& 35.4 \\
					
			CBST \cite{Zou_2018_ECCV} & & 69.6 & 28.7 &69.5& \textbf{12.1} & 0.1& 25.4& 11.9& 13.6& 82.0& \textbf{81.9}& 49.1& 14.5& 66.0& 6.6& 3.7& 32.4& 35.4 \\ 
			
			ADV \cite{Vu_2019_CVPR} &~~~\begin{rotate}{90} VGG
			\end{rotate} & 67.9& 29.4& 71.9& 6.3& 0.3& 19.9& 0.6& 2.6& 74.9& 74.9& 35.4& 9.6& 67.8& 21.4& 4.1& 15.5 &31.4 \\
			
			TGCF \cite{Choi_2019_ICCV} & & \textbf{90.1} &\textbf{48.6}& \textbf{80.7}& 2.2 &0.2& 27.2& 3.2& 14.3& \textbf{82.1}& 78.4& \textbf{54.4} &16.4 &\textbf{82.5}& 12.3& 1.7& 21.8 &38.5 \\
			
			DPR \cite{Tsai_2019_ICCV} & & 72.6& 29.5& 77.2& 3.5 &0.4 &21.0 &1.4& 7.9& 73.3& 79.0& 45.7 &14.5& 69.4& 19.6& 7.4& 16.5& 33.7 \\
			
			PyCDA \cite{Lian_2019_ICCV} & & 80.6& 26.6 &74.5& 2.0 &0.1& 18.1& \textbf{13.7}& 14.2& 80.8& 71.0& 48.0& 19.0& 72.3& 22.5& 12.1 &18.1& 35.9 \\

			\rowcolor{lightgray}
			Ours-w/oTC & & 79.4 & 36.9 & 76.1 & 5.3 & 0.6 & 25.1 & 12.4 & 12.5 & 76.7 & 78.5 & 44.6 & 22.4 & 72.8 & 28.6 & 12.5 & 25.7 & 38.1  \\
			
			\rowcolor{lightgray}
			Ours-w/oCG & & 75.0 & 32.1 & 78.2 & 2.7 & \textbf{0.8} & 22.0 & 4.2 & 7.4 & 74.3 & 81.1 & 40.7 & 23.6 & 72.3 & 34.0 & 15.9 & 27.6 & 37.0  \\
			
			\rowcolor{lightgray}
			Ours-w/oSD & & 71.2 & 30.4 & 75.7 & 6.3 & 0.6 & 26.7 & 15.6 & 17.4 & 77.1 & 80.6 & 50.3 & 21.4 & 72.0 & 29.6 & 16.7 & 26.3 & 38.6 \\
			
			\rowcolor{lightgray}
			Ours & & 70.9 & 30.5 & 77.8 & 9.0 & 0.6 & \textbf{27.3} & 8.8 & 12.9 & 74.8 & 81.1 & 43.0 & \textbf{25.1} & 73.4 & \textbf{34.5} & \textbf{19.5} & 38.2 & \textbf{39.2} \\

			\hline
			\hline
			Source only \cite{net:resnet} &  & 55.6 & 23.8 & 74.6 & 9.2 & 0.2 & 24.4 & 6.1 & 12.1 & 74.8 & 79.0 & 55.3 & 19.1 & 39.6 & 23.3 & 13.7 & 25.0 & 33.5  \\				
			
			CGAN \cite{exp:CGAN} & & 85.0 & 25.8 & 73.5 & 3.4 & \textbf{3.0} & 31.5 &19.5 & 21.3 & 67.4 & 69.4 & \textbf{68.5} & 25.0 & 76.5 & 41.6 & 17.9 & 29.5 & 41.2 \\ 
			
			DCAN \cite{Wu_2018_ECCV} &  & 81.5& 33.4& 72.4& 7.9& 0.2& 20.0& 8.6& 10.5& 71.0& 68.7& 51.5& 18.7& 75.3& 22.7& 12.8& 28.1& 36.5  \\
			
			CBST \cite{Zou_2018_ECCV} & & 53.6 & 23.7 & 75.0 & 12.5 & 0.3 & \textbf{36.4} & 23.5 & 26.3 & \textbf{84.8} & 74.7 & 67.2 & 17.5 & 84.5 & 28.4 & 15.2 & \textbf{55.8} & 42.5  \\
			
			ADV \cite{Vu_2019_CVPR} & & \textbf{85.6} & \textbf{42.2} & 79.7 & 8.7 & 0.4 & 25.9 & 5.4 & 8.1 & 80.4 & 84.1 & 57.9 & 23.8 & 73.3 & 36.4 & 14.2 & 33.0 & 41.2  \\
			
			SWLS \cite{Dong_2019_ICCV} & & 68.4 & 30.1 & 74.2 & 21.5 & 0.4 & 29.2 & 29.3 & 25.1 & 80.3 & 81.5 & 63.1 & 16.4 & 75.6 & 13.5 & 26.1 & 51.9 & 42.9  \\
			
			MSL \cite{Chen_2019_ICCV} & & 82.9& 40.7& 80.3& 10.2& 0.8& 25.8& 12.8& 18.2& 82.5& 82.2& 53.1& 18.0& 79.0& 31.4& 10.4& 35.6 &41.4 \\
			
			DPR \cite{Tsai_2019_ICCV} &~~~\begin{rotate}{90} ResNet
			\end{rotate} & 82.4 & 38.0 & 78.6 & 8.7 & 0.6 & 26.0 & 3.9 & 11.1 & 75.5 & 84.6 & 53.5 & 21.6 & 71.4 & 32.6 & 19.3 & 31.7 & 40.0 \\

			PyCDA \cite{Lian_2019_ICCV} & & 75.5 & 30.9 & \textbf{83.3} & 20.8 & 0.7 & 32.7 & 27.3 & \textbf{33.5} & 84.7 &\textbf{85.0} & 64.1 & 25.4 & 85.0 & \textbf{45.2} & 21.2 & 32.0 & 46.7  \\

			\rowcolor{lightgray}
			Ours-w/oTC & & 71.8 & 32.4 & 80.5 & 22.6 & 0.4 & 28.6 & 29.4 & 27.9 & 83.1 & 83.7 & 65.5 & 19.8 & 84.5 & 25.6 & 24.3 & 46.1 & 45.4  \\ 
			
			\rowcolor{lightgray}
			Ours-w/oCG & & 74.5 & 33.4 & 78.2 & \textbf{24.1} & 0.7 & 30.8 & \textbf{31.7} & 25.2 & 76.5 & 81.4 & 58.7 & 25.6 & 75.7 & 24.8 & 25.3 & 41.6 & 44.3  \\
			
			\rowcolor{lightgray}
			Ours-w/oSD & & 79.2 & 38.1 & 79.8 & 21.3 & 0.8 & 27.6 & 31.0 & 24.3 & 81.5 & 81.7 & 62.4 & 22.1 & \textbf{86.4} & 31.5 & 23.8 & 44.2 & 46.0 \\
			
			\rowcolor{lightgray}
			Ours & & 80.2 & 41.1 & 78.9 & 23.6 & 0.6 & 31.0 & 27.1 & 29.5 & 82.5 & 83.2 & 62.1 & \textbf{26.8} & 81.5 & 37.2 & \textbf{27.3} & 42.9 & \textbf{47.2} \\
			\hline		
		\end{tabular}
	}		
	\label{tab:Exp_SYN2City}
\end{table*}

\subsection{Experiments on SYNTHIA $\rightarrow$ Cityscapes Task}
For the experimental configurations, we regard SYNTHIA \cite{Ros_2016_CVPR} and the training subset of Cityscape \cite{Cordts_2016_CVPR} as source and target domains, and utilize the validation subset of Cityscapes to evaluate the transfer performance. From the presented comparison results in Table~\ref{tab:Exp_SYN2City}, we can notice that: 1) Compared with other competing methods such as \cite{Lian_2019_ICCV, Tsai_2019_ICCV, Chen_2019_ICCV, Dong_2019_ICCV, Vu_2019_CVPR}, our model improves the performance about 0.5\%$\sim$19.0\% mIoU to bridge both domain-wise and category-wise distributions shifts across domains. 2) Ablation studies verify that each component is indispensable to boost semantic knowledge transfer. 3) A critical transfer based adversarial framework efficiently highlights transferable domain-wise knowledge while neglecting irrelevant knowledge.

\begin{table*}[t]
	\centering
	\setlength{\tabcolsep}{1.33mm}
	\caption{Performance of our model with ResNet-101 on Cityscapes \cite{Cordts_2016_CVPR} $\rightarrow$ NTHU \cite{exp:CCA}.} 
	\scalebox{0.66}{
		\begin{tabular}{|c|c|ccccccccccccc|c|}
			\hline
			City & Method & road & sidewalk & building &light &sign & veg & sky & person & rider & car & bus & mbike & bike & mIoU \\
			\hline	
			\hline
			
			& Source only \cite{net:resnet} & 83.9& 34.3& 87.7& 13.0& 41.9& 84.6& 92.5& 37.7& 22.4& 80.8& 38.1& 39.1& 5.3& 50.9 \\
			& NMD \cite{exp:CCA} & 79.5& 29.3& 84.5& 0.0& 22.2& 80.6& 82.8& 29.5& 13.0& 71.7& 37.5& 25.9& 1.0& 42.9 \\ 
			& CBST \cite{Zou_2018_ECCV} & \cellcolor{lightgray}\textbf{87.1}& \cellcolor{lightgray}\textbf{43.9}& 89.7& 14.8& \cellcolor{lightgray}\textbf{47.7}& 85.4& 90.3& 45.4& 26.6&\cellcolor{lightgray}\textbf{85.4} &20.5& 49.8& \cellcolor{lightgray}\textbf{10.3} & 53.6  \\
			Rome & LtA \cite{exp:LtA} & 83.9& 34.2& 88.3& 18.8& 40.2&\cellcolor{lightgray}\textbf{86.2}& 93.1& 47.8& 21.7& 80.9& 47.8& 48.3& 8.6& 53.8 \\ 
			& MSL \cite{Chen_2019_ICCV}&  82.9& 32.6& 86.7& 20.7& 41.6& 85.0& 93.0& 47.2& 22.5& 82.2& \cellcolor{lightgray}\textbf{53.8}& 50.5& 9.9& 54.5 \\
			& SSF \cite{Du_2019_ICCV} & 84.2 & 38.4& 87.4 &\cellcolor{lightgray}\textbf{23.4} &43.0& 85.6& 88.2&\cellcolor{lightgray}\textbf{50.2}& 23.7& 80.6& 38.1& 51.6& 8.6& 54.1 \\
			& Ours & 85.7 & 36.5 & \cellcolor{lightgray}\textbf{92.1} & 19.4 & 42.6 & 84.8 & \cellcolor{lightgray}\textbf{95.0} & 46.9 & \cellcolor{lightgray}\textbf{28.3} & 79.4 & 40.5 & \cellcolor{lightgray}\textbf{54.2} & 7.5 & \cellcolor{lightgray}\textbf{54.8} \\
			\hline 
			
			& Source only \cite{net:resnet} & 76.6 & 47.3 & 82.5 & 12.6 & 22.5 & 77.9 & 86.5 & 43.0 & 19.8 & 74.5 & 36.8 & 29.4 & 16.7 & 48.2 \\
			& NMD \cite{exp:CCA} & 74.2& 43.9& 79.0& 2.4 &7.5& 77.8& 69.5& 39.3& 10.3& 67.9& \cellcolor{lightgray}\textbf{41.2} & 27.9& 10.9 &42.5 \\ 
			& CBST \cite{Zou_2018_ECCV} & \cellcolor{lightgray}\textbf{84.3} & \cellcolor{lightgray}\textbf{55.2}& \cellcolor{lightgray}\textbf{85.4}& \cellcolor{lightgray}\textbf{19.6}& \cellcolor{lightgray}\textbf{30.1}& 80.5& 77.9& 55.2& 28.6& \cellcolor{lightgray}\textbf{79.7}& 33.2& 37.6& 11.5& 52.2  \\
			Rio & LtA \cite{exp:LtA} & 76.2& 44.7& 84.6& 9.3& 25.5& \cellcolor{lightgray}\textbf{81.8}& 87.3& 55.3& 32.7& 74.3& 28.9& 43.0& 27.6& 51.6 \\ 
			& MSL \cite{Chen_2019_ICCV} & 76.9& 48.8& 85.2& 13.8& 18.9& 81.7& \cellcolor{lightgray}\textbf{88.1}& 54.9& 34.0& 76.8& 39.8& 44.1& 29.7& 53.3 \\
			& SSF \cite{Du_2019_ICCV} & 74.2& 43.7& 82.5& 10.3& 21.7& 79.4& 86.7& 55.9& 36.1& 74.9& 33.7& \cellcolor{lightgray}\textbf{52.6}& 33.7& 52.7 \\
			& Ours & 79.5 & 52.7 & 83.6 & 12.4 & 23.0 & 80.9 & 79.7 &\cellcolor{lightgray}\textbf{56.1} & \cellcolor{lightgray}\textbf{37.7} & 72.4 & 36.0 & 51.6 & \cellcolor{lightgray}\textbf{34.1} & \cellcolor{lightgray}\textbf{53.8} \\ 
			\hline 
			
			& Source only \cite{net:resnet} & 82.9 &31.3& 78.7& 14.2& 24.5& 81.6& 89.2& 48.6& 33.3& 70.5& 7.7& 11.5& 45.9& 47.7  \\
			& NMD \cite{exp:CCA} & 83.4& 35.4& 72.8& 12.3& 12.7& 77.4& 64.3& 42.7& 21.5& 64.1& \cellcolor{lightgray}\textbf{20.8}& 8.9& 40.3& 42.8 \\ 
			& CBST \cite{Zou_2018_ECCV} & \cellcolor{lightgray}\textbf{85.2} & 33.6& 80.4& 8.3& \cellcolor{lightgray}\textbf{31.1}& \cellcolor{lightgray}\textbf{83.9}& 78.2 &53.2& 28.9& \cellcolor{lightgray}\textbf{72.7}& 4.4& 27.0& 47.0& 48.8  \\
			Tokyo & LtA \cite{exp:LtA} & 81.5& 26.0& 77.8& 17.8& 26.8& 82.7& \cellcolor{lightgray}\textbf{90.9}& 55.8& 38.0& 72.1& 4.2& 24.5& 50.8& 49.9 \\ 
			& MSL \cite{Chen_2019_ICCV} & 81.2& 30.1& 77.0& 12.3& 27.3& 82.8& 89.5& 58.2& 32.7& 71.5& 5.5& \cellcolor{lightgray}\textbf{37.4}& 48.9& 50.5 \\
			& SSF \cite{Du_2019_ICCV} & 82.1& 27.4& 78.0& \cellcolor{lightgray}\textbf{18.4}& 26.6& 83.0& 90.8& 57.1& 35.8& 72.0& 4.6& 27.3& \cellcolor{lightgray}\textbf{52.8} & 50.4 \\
			& Ours & 83.1 & \cellcolor{lightgray}\textbf{35.5} & \cellcolor{lightgray}\textbf{81.2} & 15.8 & 24.9 & 81.3 & 86.4 & \cellcolor{lightgray}\textbf{58.8} & \cellcolor{lightgray}\textbf{39.2} & 68.1 & 6.7 & 30.4 & 51.2 & \cellcolor{lightgray}\textbf{51.0}  \\
			\hline 
			
			& Source only \cite{net:resnet} & 83.5& 33.4& 86.6& 12.7 &16.4& 77.0& \cellcolor{lightgray}\textbf{92.1}& 17.6& 13.7& 70.7& 37.7& 44.4& 18.5& 46.5 \\
			& NMD \cite{exp:CCA} & 78.6& 28.6& 80.0& 13.1& 7.6& 68.2& 82.1& 16.8& 9.4& 60.4& 34.0& 26.5& 9.9& 39.6 \\ 
			& CBST \cite{Zou_2018_ECCV} & \cellcolor{lightgray}\textbf{86.1}& 35.2& 84.2& 15.0& \cellcolor{lightgray}\textbf{22.2}& 75.6& 74.9& 22.7& \cellcolor{lightgray}\textbf{33.1}& \cellcolor{lightgray}\textbf{78.0}& 37.6& \cellcolor{lightgray}\textbf{58.0}& 30.9& 50.3  \\
			Taipei & LtA \cite{exp:LtA} & 81.7 & 29.5 & 85.2& 26.4& 15.6& 76.7& 91.7& 31.0& 12.5& 71.5& 41.1& 47.3& 27.7& 49.1 \\ 
			& MSL \cite{Chen_2019_ICCV}& 80.7& 32.5& 85.5& \cellcolor{lightgray}\textbf{32.7} & 15.1& \cellcolor{lightgray}\textbf{78.1}& 91.3& 32.9& 7.6& 69.5& \cellcolor{lightgray}\textbf{44.8} &52.4& \cellcolor{lightgray}\textbf{34.9}& 50.6 \\
			& SSF \cite{Du_2019_ICCV} & 84.5 &\cellcolor{lightgray}\textbf{35.3}& 86.4& 17.7& 16.9& 77.7& 91.3& 31.8& 22.3& 73.7& 41.1& 55.9& 28.5& 51.0 \\
			& Ours & 83.4 & 33.7 & \cellcolor{lightgray}\textbf{87.5} & 24.3 & 17.2 & 75.8 & 90.6 & \cellcolor{lightgray}\textbf{33.2} & 24.1 & 75.3 & 35.8 & 56.4 & 31.2 & \cellcolor{lightgray}\textbf{51.4}   \\
			\hline 
			
		\end{tabular}
	}		
	\label{tab:Exp_CrossCity}
\end{table*}

\begin{figure*}[h]
	\small
	\centering
	\includegraphics[trim = 5.5mm 112mm 25mm 140mm, clip, width =345pt, height =90pt]{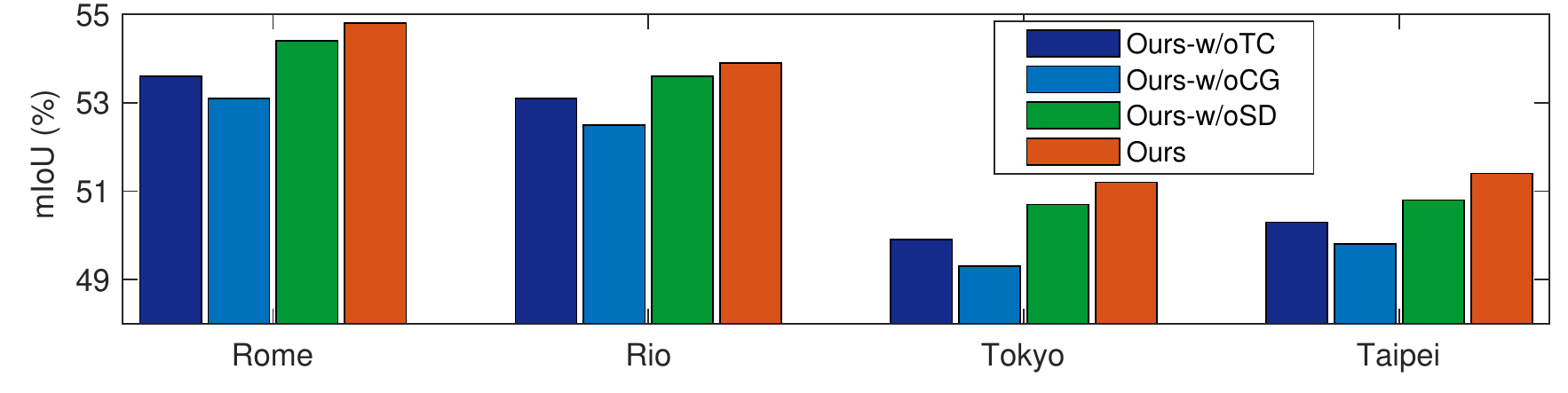}
	\caption{Ablation studies of our model with ResNet-101 on Cityscapes $\rightarrow$ NTHU task.}  
	\label{fig:NTHU_ablation_studies} 
\end{figure*}

\subsection{Experiments on Cityscapes $\rightarrow$ NTHU Task}
As for Cityscapes $\rightarrow$ NTHU task, we respectively regard the training subset of Cityscapes and NTHU as source and target domains. The testing subset of every city in NTHU is utilized for evaluation. Table~\ref{tab:Exp_CrossCity} reports the comparison adaptation results, and the corresponding ablation studies are depicted in Fig.~\ref{fig:NTHU_ablation_studies}. Some conclusions are drawn from Table~\ref{tab:Exp_CrossCity} and Fig.~\ref{fig:NTHU_ablation_studies}: 
1) Our model achieves about 0.3\% $\sim$ 11.9\% improvement than other methods across all the evaluated cities. 2) Ablation experiments validate the rationality and effectiveness of each module. 3) With the guidance of confident soft pseudo labels, $\mathcal{L}_{div}$ could map the semantic features of same class across domains compactly nearby.

\section{Conclusion}
In this paper, we propose a new Critical Semantic-Consistent Learning (CSCL) model, which aims to narrow both domain-wise and category-wise distributions shifts. A critical transfer based adversarial framework is developed to highlight transferable domain-wise knowledge while preventing the negative transfer of irrelevant knowledge. Specifically, transferability-critic feedbacks positive guidance to transferability-quantizer when negative transfer occurs. Moreover, with confidence-guided pseudo label generator assigning soft pseudo labels for target samples, a symmetric soft divergence loss is designed to minimize category-wise discrepancy. Experiments on public datasets verify the effectiveness of our model.

\section*{Acknowledgment} 
This work is supported by Ministry of Science and Technology of the People´s Republic of China (2019YFB1310300), National Nature Science Foundation of China under Grant (61722311, U1613214, 61821005, 61533015) and National Postdoctoral Innovative Talents Support Program (BX20200353).

\clearpage
%
%
\bibliographystyle{splncs04}
\bibliography{0696}

\begin{thebibliography}{10}
\providecommand{\url}[1]{\texttt{#1}}
\providecommand{\urlprefix}{URL }
\providecommand{\doi}[1]{https://doi.org/#1}

\bibitem{Chen_Semantic2014}
Chen, L.C., Papandreou, G., Kokkinos, I., Murphy, K., Yuille, A.: Semantic
  image segmentation with deep convolutional nets and fully connected crfs.
  arXiv preprint arXiv:1412.7062  (December 2014)

\bibitem{net:deeplab}
Chen, L.C., Zhu, Y., Papandreou, G., Schroff, F., Adam, H.: Encoder-decoder
  with atrous separable convolution for semantic image segmentation. In: The
  European Conference on Computer Vision (ECCV) (September 2018)

\bibitem{Chen_2019_ICCV}
Chen, M., Xue, H., Cai, D.: Domain adaptation for semantic segmentation with
  maximum squares loss. In: The IEEE International Conference on Computer
  Vision (ICCV) (October 2019)

\bibitem{exp:CCA}
Chen, Y.H., Chen, W.Y., Chen, Y.T., Tsai, B.C., Frank~Wang, Y.C., Sun, M.: No
  more discrimination: Cross city adaptation of road scene segmenters. In: The
  IEEE International Conference on Computer Vision (ICCV) (Oct 2017)

\bibitem{Choi_2019_ICCV}
Choi, J., Kim, T., Kim, C.: Self-ensembling with gan-based data augmentation
  for domain adaptation in semantic segmentation. In: The IEEE International
  Conference on Computer Vision (ICCV) (October 2019)

\bibitem{Cordts_2016_CVPR}
Cordts, M., Omran, M., Ramos, S., Rehfeld, T., Enzweiler, M., Benenson, R.,
  Franke, U., Roth, S., Schiele, B.: The cityscapes dataset for semantic urban
  scene understanding. In: The IEEE Conference on Computer Vision and Pattern
  Recognition (CVPR) (June 2016)

\bibitem{NIPS2017_6963}
Courty, N., Flamary, R., Habrard, A., Rakotomamonjy, A.: Joint distribution
  optimal transportation for domain adaptation. In: Guyon, I., Luxburg, U.V.,
  Bengio, S., Wallach, H., Fergus, R., Vishwanathan, S., Garnett, R. (eds.)
  Advances in Neural Information Processing Systems 30, pp. 3730--3739. Curran
  Associates, Inc. (2017)

\bibitem{5206848}
{Deng}, J., {Dong}, W., {Socher}, R., {Li}, L., {Kai Li}, {Li Fei-Fei}:
  Imagenet: A large-scale hierarchical image database. In: 2009 IEEE Conference
  on Computer Vision and Pattern Recognition. pp. 248--255 (June 2009)

\bibitem{Ding_2018_ECCV}
Ding, Z., Li, S., Shao, M., Fu, Y.: Graph adaptive knowledge transfer for
  unsupervised domain adaptation. In: Proceedings of the European Conference on
  Computer Vision (ECCV) (September 2018)

\bibitem{Dong_2019_ICCV}
Dong, J., Cong, Y., Sun, G., Hou, D.: Semantic-transferable weakly-supervised
  endoscopic lesions segmentation. In: The IEEE International Conference on
  Computer Vision (ICCV) (October 2019)

\bibitem{Dong_2020_CVPR}
Dong, J., Cong, Y., Sun, G., Zhong, B., Xu, X.: What can be transferred:
  Unsupervised domain adaptation for endoscopic lesions segmentation. In:
  IEEE/CVF Conference on Computer Vision and Pattern Recognition (CVPR) (June
  2020)

\bibitem{Du_2019_ICCV}
Du, L., Tan, J., Yang, H., Feng, J., Xue, X., Zheng, Q., Ye, X., Zhang, X.:
  Ssf-dan: Separated semantic feature based domain adaptation network for
  semantic segmentation. In: The IEEE International Conference on Computer
  Vision (ICCV) (October 2019)

\bibitem{Gong_2019_CVPR}
Gong, R., Li, W., Chen, Y., Gool, L.V.: Dlow: Domain flow for adaptation and
  generalization. In: The IEEE Conference on Computer Vision and Pattern
  Recognition (CVPR) (June 2019)

\bibitem{Goodfellow:2014:GAN}
Goodfellow, I.J., Pouget-Abadie, J., Mirza, M., Xu, B., Warde-Farley, D.,
  Ozair, S., Courville, A., Bengio, Y.: Generative adversarial nets. In:
  Proceedings of the 27th International Conference on Neural Information
  Processing Systems - Volume 2. pp. 2672--2680 (2014)

\bibitem{net:resnet}
He, K., Zhang, X., Ren, S., Sun, J.: Deep residual learning for image
  recognition. In: The IEEE Conference on Computer Vision and Pattern
  Recognition (CVPR) (June 2016)

\bibitem{exp:Wild}
Hoffman, J., Wang, D., Yu, F., Darrell, T.: Fcns in the wild: Pixel-level
  adversarial and constraint-based adaptation. arXiv preprint arXiv:1612.02649
  (2016)

\bibitem{exp:CGAN}
Hong, W., Wang, Z., Yang, M., Yuan, J.: Conditional generative adversarial
  network for structured domain adaptation. In: The IEEE Conference on Computer
  Vision and Pattern Recognition (CVPR) (June 2018)

\bibitem{Lee_2019_CVPR}
Lee, C.Y., Batra, T., Baig, M.H., Ulbricht, D.: Sliced wasserstein discrepancy
  for unsupervised domain adaptation. In: The IEEE Conference on Computer
  Vision and Pattern Recognition (CVPR) (June 2019)

\bibitem{Lee_2019_ICCV}
Lee, S., Kim, D., Kim, N., Jeong, S.G.: Drop to adapt: Learning discriminative
  features for unsupervised domain adaptation. In: The IEEE International
  Conference on Computer Vision (ICCV) (October 2019)

\bibitem{Li_2019_CVPR}
Li, Y., Yuan, L., Vasconcelos, N.: Bidirectional learning for domain adaptation
  of semantic segmentation. In: The IEEE Conference on Computer Vision and
  Pattern Recognition (CVPR) (June 2019)

\bibitem{Lian_2019_ICCV}
Lian, Q., Lv, F., Duan, L., Gong, B.: Constructing self-motivated pyramid
  curriculums for cross-domain semantic segmentation: A non-adversarial
  approach. In: The IEEE International Conference on Computer Vision (ICCV)
  (October 2019)

\bibitem{NIPS2017_6750}
Liu, S., De~Mello, S., Gu, J., Zhong, G., Yang, M.H., Kautz, J.: Learning
  affinity via spatial propagation networks. In: Guyon, I., Luxburg, U.V.,
  Bengio, S., Wallach, H., Fergus, R., Vishwanathan, S., Garnett, R. (eds.)
  Advances in Neural Information Processing Systems 30, pp. 1520--1530. Curran
  Associates, Inc. (2017)

\bibitem{Liu_2015_ICCV}
Liu, Z., Li, X., Luo, P., Loy, C.C., Tang, X.: Semantic image segmentation via
  deep parsing network. In: The IEEE International Conference on Computer
  Vision (ICCV) (December 2015)

\bibitem{Luo_2019_ICCV}
Luo, Y., Liu, P., Guan, T., Yu, J., Yang, Y.: Significance-aware information
  bottleneck for domain adaptive semantic segmentation. In: The IEEE
  International Conference on Computer Vision (ICCV) (October 2019)

\bibitem{Luo_2019_CVPR}
Luo, Y., Zheng, L., Guan, T., Yu, J., Yang, Y.: Taking a closer look at domain
  shift: Category-level adversaries for semantics consistent domain adaptation.
  In: The IEEE Conference on Computer Vision and Pattern Recognition (CVPR)
  (June 2019)

\bibitem{Richter_2016_ECCV}
Richter, S.R., Vineet, V., Roth, S., Koltun, V.: Playing for data: {G}round
  truth from computer games. In: Leibe, B., Matas, J., Sebe, N., Welling, M.
  (eds.) European Conference on Computer Vision (ECCV). LNCS, vol.~9906, pp.
  102--118. Springer International Publishing (2016)

\bibitem{RonnebergerFB15}
Ronneberger, O., Fischer, P., Brox, T.: U-net: Convolutional networks for
  biomedical image segmentation. arXiv preprint arXiv:1505.04597  (August 2015)

\bibitem{Ros_2016_CVPR}
Ros, G., Sellart, L., Materzynska, J., Vazquez, D., Lopez, A.M.: The synthia
  dataset: A large collection of synthetic images for semantic segmentation of
  urban scenes. In: The IEEE Conference on Computer Vision and Pattern
  Recognition (CVPR) (June 2016)

\bibitem{Saito_2018_CVPR}
Saito, K., Watanabe, K., Ushiku, Y., Harada, T.: Maximum classifier discrepancy
  for unsupervised domain adaptation. In: The IEEE Conference on Computer
  Vision and Pattern Recognition (CVPR) (June 2018)

\bibitem{exp:LSD}
Sankaranarayanan, S., Balaji, Y., Jain, A., Nam~Lim, S., Chellappa, R.:
  Learning from synthetic data: Addressing domain shift for semantic
  segmentation. In: The IEEE Conference on Computer Vision and Pattern
  Recognition (CVPR) (June 2018)

\bibitem{Shelhamer:2017:FCN:3069214.3069246}
Shelhamer, E., Long, J., Darrell, T.: Fully convolutional networks for semantic
  segmentation. IEEE Trans. Pattern Anal. Mach. Intell.  \textbf{39}(4),
  640--651 (Apr 2017)

\bibitem{Simonyan15}
Simonyan, K., Zisserman, A.: Very deep convolutional networks for large-scale
  image recognition. In: International Conference on Learning Representations
  (2015)

\bibitem{exp:LtA}
Tsai, Y.H., Hung, W.C., Schulter, S., Sohn, K., Yang, M.H., Chandraker, M.:
  Learning to adapt structured output space for semantic segmentation. In: The
  IEEE Conference on Computer Vision and Pattern Recognition (CVPR) (June 2018)

\bibitem{Tsai_2019_ICCV}
Tsai, Y., Sohn, K., Schulter, S., Chandraker, M.: Domain adaptation for
  structured output via discriminative patch representations. In: The IEEE
  International Conference on Computer Vision (ICCV) (October 2019)

\bibitem{Vu_2019_CVPR}
Vu, T.H., Jain, H., Bucher, M., Cord, M., Perez, P.: Advent: Adversarial
  entropy minimization for domain adaptation in semantic segmentation. In: The
  IEEE Conference on Computer Vision and Pattern Recognition (CVPR) (June 2019)

\bibitem{wang2019laplacian}
Wang, Q., Fan, H., Sun, G., Cong, Y., Tang, Y.: Laplacian pyramid adversarial
  network for face completion. Pattern Recognition  \textbf{88},  493--505
  (2019)

\bibitem{wang2020recurrent}
Wang, Q., Fan, H., Sun, G., Ren, W., Tang, Y.: Recurrent generative adversarial
  network for face completion. IEEE Transactions on Multimedia  (2020)

\bibitem{Wu_2018_ECCV}
Wu, Z., Han, X., Lin, Y.L., Gokhan~Uzunbas, M., Goldstein, T., Nam~Lim, S.,
  Davis, L.S.: Dcan: Dual channel-wise alignment networks for unsupervised
  scene adaptation. In: The European Conference on Computer Vision (ECCV)
  (September 2018)

\bibitem{Xia_2020_CVPR}
Xia, H., Ding, Z.: Structure preserving generative cross-domain learning. In:
  IEEE/CVF Conference on Computer Vision and Pattern Recognition (CVPR) (June
  2020)

\bibitem{Yu2015MultiScaleCA}
Yu, F., Koltun, V.: Multi-scale context aggregation by dilated convolutions.
  arXiv preprint arXiv:1511.07122  (November 2015)

\bibitem{10.5555/3042817.3043028}
Zhang, K., Sch\"{o}lkopf, B., Muandet, K., Wang, Z.: Domain adaptation under
  target and conditional shift. In: Proceedings of the 30th International
  Conference on International Conference on Machine Learning - Volume 28. p.
  III–819–III–827. ICML’13, JMLR.org (2013)

\bibitem{zhang2019visual}
Zhang, T., Cong, Y., Sun, G., Wang, Q., Ding, Z.: Visual tactile fusion object
  clustering. In: AAAI Conference on Artificial Intelligence (2020)

\bibitem{exp:CL}
Zhang, Y., David, P., Gong, B.: Curriculum domain adaptation for semantic
  segmentation of urban scenes. In: The IEEE International Conference on
  Computer Vision (ICCV) (Oct 2017)

\bibitem{Zhao_2018_ECCV}
Zhao, H., Zhang, Y., Liu, S., Shi, J., Change~Loy, C., Lin, D., Jia, J.:
  Psanet: Point-wise spatial attention network for scene parsing. In: The
  European Conference on Computer Vision (ECCV) (September 2018)

\bibitem{Zhu_2019_ICCV}
Zhu, Z., Xu, M., Bai, S., Huang, T., Bai, X.: Asymmetric non-local neural
  networks for semantic segmentation. In: The IEEE International Conference on
  Computer Vision (ICCV) (October 2019)

\bibitem{Zou_2019_ICCV}
Zou, Y., Yu, Z., Liu, X., Kumar, B.V., Wang, J.: Confidence regularized
  self-training. In: Proceedings of the IEEE/CVF International Conference on
  Computer Vision (ICCV) (October 2019)

\bibitem{Zou_2018_ECCV}
Zou, Y., Yu, Z., Vijaya~Kumar, B., Wang, J.: Unsupervised domain adaptation for
  semantic segmentation via class-balanced self-training. In: The European
  Conference on Computer Vision (ECCV) (September 2018)

\end{thebibliography}
\end{document}